\documentclass{article}

\usepackage[preprint]{neurips_2026}

\usepackage{microtype}
\usepackage{graphicx}
\usepackage{subfigure}
\usepackage{subcaption}
\usepackage{booktabs} % for professional tables
\usepackage[table]{xcolor}
\usepackage{multirow} 
\usepackage{overpic}
\usepackage{float}
\usepackage{algorithm}
\usepackage{algorithmic}

\usepackage{hyperref}
\usepackage{enumitem}
\usepackage{amsmath}
\usepackage{wrapfig}
\usepackage{mdframed}

\usepackage{tikz}
\usetikzlibrary{shapes.geometric, arrows.meta, positioning, fit, calc}

\usepackage[utf8]{inputenc} % allow utf-8 input
\usepackage[T1]{fontenc}    % use 8-bit T1 fonts
\usepackage{hyperref}       % hyperlinks
\usepackage{url}            % simple URL typesetting
\usepackage{booktabs}       % professional-quality tables
\usepackage{amsfonts}       % blackboard math symbols
\usepackage{nicefrac}       % compact symbols for 1/2, etc.
\usepackage{microtype}      % microtypography
\usepackage{xcolor}         % colors

\title{Layerwise Progressive Freezing: A Training Scaffold for Depth-Scalable Binary Networks}

\author{%
  Evan Gibson Smith \\
  Worcester Polytechnic Institute\\
  \texttt{egsmith@wpi.edu} 
  \And
  Bashima Islam \\
  Worcester Polytechnic Institute\\
  \texttt{bislam@wpi.edu} \\
}

\begin{document}

\maketitle

\begin{abstract}
Training binary neural networks (BNNs) from scratch is dominated by the straight-through estimator (STE), whose forward/backward mismatch produces severe accuracy degradation as networks deepen. We study an orthogonal axis: when and where binarization is enforced during training. We introduce StoMPP (Stochastic Masked Partial Progressive Binarization), which gradually replaces clipped weights and activations with their hard binary counterparts layer by layer from input to output, using stochastic partial masks with soft refresh. StoMPP delivers two complementary benefits. As a standalone training rule, it provides a fully STE-free procedure that improves over vanilla STE with gains that grow with depth (ResNet-50 BNN: +18.0/+13.5/+3.8 on CIFAR-10/100/ImageNet), and the pattern holds across ResNet-18/34/50, MobileNetV2, and BERT fine-tuning. Composed with surrogate gradients by applying STE only to frozen entries, it reaches +27.1/+19.8/+17.7 over vanilla STE on the same setting. Underlying both regimes is a single mechanistic finding: progression order is decisive. Forward layerwise progression prevents depth collapse, reverse progression collapses to near-chance, and binary-weight networks (without binary activations) are insensitive to order. We trace this asymmetry to activation-induced gradient blockades: a committed binary activation severs gradient flow upstream, and ordering controls when these blockades form. To isolate the progression's contribution from any benefit conferred by STE, we conduct all ablations in the STE-free regime; the resulting characterization (schedule, refresh, ordering, dynamics) thus reflects the progression itself rather than its interaction with surrogate gradients.
\end{abstract}

\section{Introduction}

Deep neural networks have made dramatic progress on tasks ranging from vision to language, but their growing memory and compute footprints make deployment on resource-constrained hardware increasingly difficult. Binary neural networks address this by constraining weights, and sometimes activations, to $\{-1,+1\}$, which can reduce inference cost by an order of magnitude or more. These methods fall into two regimes: binary-weight networks (BWNs), where only weights are binarized, and full binary neural networks (BNNs), where activations are binarized as well. Training either regime from scratch is difficult: the forward pass uses a non-differentiable sign operator, and standard gradient-based optimization requires some way to propagate learning signal through it. The dominant solution is the straight-through estimator (STE) \citep{BinaryConnect_Courbariaux_2016, ste_bengio_2013}, which substitutes a surrogate gradient in the backward pass. STE has enabled most successful BNN training pipelines, but its forward/backward mismatch is associated with progressive accuracy degradation as networks deepen.

This depth sensitivity is  known~\citep{Zhang_Li_Liu_2023}, and most prior work responds to it by improving STE itself. Approaches include refined approximations of the sign function, gradient shaping, learned scaling, and specialized optimizers~\citep{ReCU_Xu_2021, BiPer_Vargas_2024a, OvSW_Xiang, XNOR_Net_Rastegari_2016, BiReal_Liu_2018, ReActNet_Liu_2020}. These efforts retain the basic STE structure of a hard forward operator paired with a surrogate backward pass, and seek to reduce the resulting mismatch from within that structure. We pursue an orthogonal direction: rather than improving the gradient that flows through a fixed binarization, we change when and where binarization is enforced during training. Progressive quantization methods offer one realization of this idea, gradually committing parameters from continuous to discrete values over the course of training, and have been used effectively for weight-only quantization~\cite{INQ_Zhou_2017a, AlphaBlend_Liu_Mattina_2019, Bai_Wang_Liberty_2019a, Yin_Zhang_Lyu_Osher_Qi_Xin_2018, Soft_then_hard_Guo_2024, Self_Binarizing_Lahoud_2019}. Extending this strategy to full BNNs is non-trivial. Naively applying a global progressive schedule, in which any parameter or activation in the network can be committed at any time, succeeds for BWNs but fails for BNNs: training collapses or stalls at depth even though the same schedule works in the weight-only setting.

We trace this failure to a mechanism we call \textit{activation-induced gradient blockades}. In a full BNN, freezing a binary activation replaces it with a hard sign operator whose derivative is zero almost everywhere; once a unit is committed in this way, gradients cannot propagate through it to earlier layers, preventing learning in deep networks (Fig. \ref{fig:masking_comparison}). Under a global progressive schedule, frozen activations can appear at arbitrary depths, and any path through a frozen activation loses gradient signal upstream. BWNs are resistant to this failure because their activations remain continuous, so freezing only the weights leaves gradient flow intact. This asymmetry is the central observation that motivates our method.

We introduce StoMPP (\textbf{Sto}chastic \textbf{M}asked \textbf{P}artial \textbf{P}rogressive Binarization), a layerwise progressive freezing procedure with stochastic masking. StoMPP progressively hardens both weights and activations from a differentiable clipped operator to a hard binary operator, scheduling commitment from input to output so that the layer currently transitioning always has an unfrozen suffix providing a gradient path to the loss. The frozen-entry gradient is a design choice: setting it to zero yields a fully STE-free training procedure, while applying STE only to frozen entries (StoMPP+STE) recovers a learning signal through committed units. Both regimes share the same progression structure, and we ablate the progression in the STE-free regime to isolate its contribution from any benefit conferred by surrogate gradients.

Our contributions are as follows:
\begin{enumerate}
    \item We introduce StoMPP, a layerwise progressive freezing procedure with stochastic masked partial freezing and soft refresh. The progression is independent of the choice of frozen-entry gradient, supporting both a fully STE-free training procedure and a composition with STE applied only to frozen entries.
    \item We show that progression order is decisive for full BNNs: forward layerwise succeeds on deep networks, reverse layerwise collapses to near-chance, and global progression degrades at depth, while BWNs are largely insensitive to ordering. We discuss this asymmetry in terms of activation-induced gradient blockades.
    \item We analyze the progression's training dynamics, including non-monotonic sawtooth convergence as layers transition, and show that layerwise scheduling is substantially more robust to schedule shape and refresh rate than global scheduling.
\end{enumerate}

Our evaluation shows that both StoMPP variants improve over a BinaryConnect-style STE baseline under a matched minimal training recipe, with gains that grow with depth. On ResNet-50 BNNs, STE-free StoMPP improves over vanilla STE by +18.0/+13.5/+3.8 on CIFAR-10/100/ImageNet, while StoMPP+STE reaches +27.1/+19.8/+17.7 on the same setting. The same pattern holds across ResNet-18/34, MobileNetV2, and BERT fine-tuning on SST-2, indicating that the progression's benefit is not specific to ResNet-style vision models. We further show that StoMPP transfers to binary-specific architectures by combining it with Bi-Real Net without architectural modification.

\section{Related Work}

\textbf{Binary Neural Networks and STE.}
Binary neural networks (BNNs) quantize weights and activations to $\{-1,+1\}$, while binary-weight networks (BWNs) quantize only weights. The dominant approach for training BNNs is the straight-through estimator (STE) \citep{BinaryConnect_Courbariaux_2016, ste_bengio_2013}, which applies the non-differentiable $\mathrm{sign}(\cdot)$ in the forward pass and uses a surrogate gradient in the backward pass. STE enabled a wide range of successful BNNs \citep{XNOR_Net_Rastegari_2016, BiReal_Liu_2018, ReActNet_Liu_2020, OvSW_Xiang}, but it introduces an inherent forward/backward mismatch: the backward update does not correspond to the gradient of the discrete forward computation. This can result in optimization instability and degraded scaling in deeper networks under standard training recipes \citep{Zhang_Li_Liu_2023, BNN_Survey_2020}. Largely, literature improves STE training by proposing better surrogate gradients or gradient shaping to reduce the mismatch and stabilize training \citep{ReCU_Xu_2021, OvSW_Xiang, BiPer_Vargas_2024a, BiReal_Liu_2018, IR_Net_2020}. Beyond convolutional networks, binarization of transformers such as BERT \citep{Devlin_Chang_Lee_Toutanova_2019} or ViTs \citep{Dosovitskiy_Beyer_Kolesnikov_Weissenborn_Zhai_Unterthiner_Dehghani_Minderer_Heigold_Gelly_et}
\citep{Qin_Ding_Zhang_Yan_Liu_Dang_Liu_Liu_2022, Liu_Oguz_Pappu_Xiao_Yih_Li_Krishnamoorthi_Mehdad_2022, Bai_Zhang_Hou_Shang_Jin_Jiang_Liu_Lyu_King_2021}, and large language models \citep{Wang_Ma_Dong_Huang_Wang_Ma_Yang_Wang_Wu_Wei_2023, zhang2026sparsebitnet158bitllmsnaturally} are becoming very popular to deploy these costly models more effectively.

\textbf{Differentiable Relaxations and Annealing.}
To avoid STE, several works optimize differentiable relaxations that are annealed toward a hard quantizer \citep{Self_Binarizing_Lahoud_2019, Yin_Zhang_Lyu_Osher_Qi_Xin_2018, Soft_then_hard_Guo_2024}. During the relaxed phase, forward and backward computations are aligned, but training can become sensitive as the relaxation hardens. We seek to avoid this.

\textbf{Progressive Quantization and Freezing.}
Progressive quantization methods convert a continuous model to a quantized one in stages by freezing subsets of parameters and retraining the remainder. INQ \citep{INQ_Zhou_2017a} progressively fixes groups of weights to quantized values in groups deterministically so remaining weights can adapt, and related stochastic methods probabilistically freeze weights during training \citep{Stochastic_Quantization_Dong_2017}. These approaches are largely developed for weight-only quantization and do not address the additional difficulty of \emph{binary activations}, which can block gradient flow in the absence of STE. 
StoMPP extends progressive freezing to full BNNs (weights and activations) by combining (i) a layerwise binarization schedule and (ii) stochastic masking. Layerwise approaches for binary neural networks appear in BiTAT \citep{BiTAT_Park_2022}, which unlike StoMPP relies on STE and layerwise correlations.

%layerwise quantization in a \emph{pretrained finetuning} setting and relies on STE, whereas StoMPP targets end-to-end BNN training without STE.

% \textcolor{red}{TODO paragraph below I think should be cut, or greatly reduced (as blue above does) because I think it might step outside related work and we need space}

% StoMPP extends progressive freezing to full BNNs (weights and activations) by combining (i) a layerwise binarization schedule and (ii) stochastic masking that refreshes a controlled fraction of discrete variables each update, preserving learning signal while maintaining stability. Relatedly, BiTAT \citep{BiTAT_Park_2022} performs layerwise quantization in a \emph{pretrained finetuning} setting and relies on STE, whereas StoMPP targets end-to-end BNN training without STE.

\textbf{Broader Quantization Training Perspectives.}
Beyond BNNs, quantization-aware training (QAT) methods often target multi-bit quantization using differentiable approximations, learned quantizer parameters (e.g., learned step sizes or clipped activations), or proximal methods. \citep{DoReFa_Zhou_2018, LSQ_Esser_2020, PACT_Choi_2018, LQ_Net_Zhang_2018, ADMM_NN_Ren_2018, Bai_Wang_Liberty_2019a}.

% \textcolor{red}{TODO: I think all of this red here can go
% minimization and ADMM/proximal-style methods \citep{ADMM_NN_Ren_2018, Bai_Wang_Liberty_2019a}
% I think Quantization is also studied through constrained/discrete optimization lenses, including alternating minimization and ADMM/proximal-style methods \citep{ADMM_NN_Ren_2018, Bai_Wang_Liberty_2019a}. While complementary, these lines typically do not address end-to-end training with \emph{binary activations} in the \textit{absence} of STE, which is the setting we study.
% }

\textbf{Orthogonal Architectural Improvements.}
A complementary direction improves BNN accuracy via architecture changes, such as learned scaling factors \citep{XNOR_Net_Rastegari_2016}, specialized activation designs \citep{ReActNet_Liu_2020}, or modified residual connections and information pathways \citep{IR_Net_2020, BiReal_Liu_2018}. Many of these introduce additional parameters or computation to compensate for binarization \citep{XNOR_Net_Rastegari_2016, XNOR_Plusplus_Bulat_2019, ReActNet_Liu_2020}. StoMPP is orthogonal: we focus on the training procedure for discrete weights and activations without surrogate gradients, and can be combined with such architectural improvements.

Appendix~\ref{sec:app_related_work} presents further related works.
\section{Method}
\label{sec:method}
% We introduce \textbf{StoMPP} (\textbf{Sto}chastic \textbf{M}asked \textbf{P}artial \textbf{P}rogressive binarization), \textcolor{red}{both an STE-free BNN training procedure and technique for improvement with STE} which progressively enforces binarization while preserving learning via \textbf{(i)} stochastic masked partial freezing (soft refresh) and \textbf{(ii)} layerwise scheduling to avoid activation-induced gradient blockades.

\subsection{StoMPP Overview}
StoMPP maintains real-valued underlying variables for both weights and pre-activations and uses a mask to decide, for each entry, whether the variable is currently treated as discrete (frozen) or continuous (unfrozen). Over the course of training, the mask is gradually filled in from input to output; by the end of training, all scheduled variables are discrete and the network is fully binary at inference.

\textbf{Forward Mapping.}
Let $u$ denote either a weight entry or a pre-activation entry. We define the continuous proxy
$\mathrm{clip}(u)=\max(-1,\min(1,u))$
and the binary map $\mathrm{sign}(u)\in\{-1,+1\}$.
Given a binary mask $M\in\{0,1\}$ (same shape as $u$), the forward value is
\begin{equation}
\label{eq:stompp_forward}
u' \;=\; M \odot \mathrm{sign}(u) \;+\; (1-M)\odot \mathrm{Smooth}(u),
\end{equation}
where $\odot$ is elementwise multiplication.
We use $\mathrm{clip}$ and $\mathrm{identity}$ as $\mathrm{Smooth}$ for activations and weights, respectively, with $\mathrm{sign}$ to match canonical STE-style BNN parameterizations. StoMPP is independent to these choices and can pair with alternative binarizers (e.g., \cite{BiReal_Liu_2018}).

\textbf{Backward Pass.}
Frozen entries (where $M = 1$) have no local gradient through $\mathrm{sign}$, since $\partial \mathrm{sign}(u) / \partial u = 0$ almost everywhere. StoMPP treats the gradient that flows through frozen entries as a design choice, captured by a function $\mathrm{Surr}(u)$:
\begin{equation}
\nabla_{u} = \nabla_{u'} \odot \left[(1{-}M) \odot \frac{\partial \mathrm{Smooth}(u)}{\partial u} + M \odot \mathrm{Surr}(u)\right]
\label{eq:backward_ste}
\end{equation}

Two settings of $\mathrm{Surr}$ are of primary interest. STE-free StoMPP sets $\mathrm{Surr}(u) = 0$, so frozen entries receive zero gradient and learning signal flows only through unfrozen (clipped) entries; this is the variant we use throughout our ablations. StoMPP+STE sets $\mathrm{Surr}(u) = 1$, used in our main results, applying a standard STE-style identity surrogate to frozen entries while preserving the progression structure. {STE baseline has mask M = 1 everywhere, frozen-entry gradient via STE identity; StoMPP (STE-free) has progressive masking, frozen entries receiving zero gradient; and StoMPP+STE uses progressive masking, where frozen entires recieve identity (STE) gradient.

\subsection{Stochastic Masked Progressive Freezing}
This subsection describes how StoMPP evolves the mask within a single layer that is currently transitioning from continuous to binary. The cross-layer schedule is deferred to Section~\ref{sec:layerwise-scheduling}.
% This component progressively increases the fraction of frozen (binary) entries within a layer while softly refreshing which specific entries are frozen, balancing stability and exploration during training.

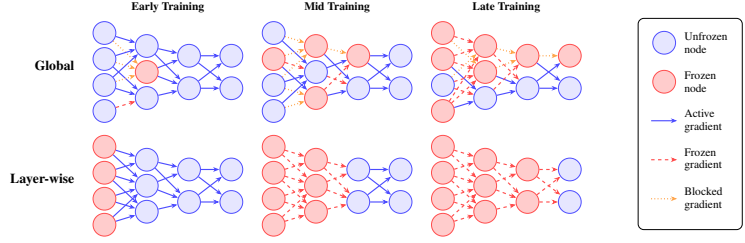
\begin{figure*}[t]
\centering

% Define styles with color
\tikzstyle{unfrozen_node} = [circle, draw=blue!70, fill=blue!10, thick, minimum size=7mm]
\tikzstyle{frozen_node} = [circle, draw=red!70, fill=red!20, thick, minimum size=7mm]
\tikzstyle{active_edge} = [-{Stealth[length=2mm]}, thick, blue!70, line width=1pt]
\tikzstyle{frozen_edge} = [-{Stealth[length=2mm]}, thick, red!70, dashed, line width=1pt]
\tikzstyle{blocked_edge} = [-{Stealth[length=2mm]}, thick, orange!70, dotted, line width=1.2pt]

\resizebox{0.7\textwidth}{!}{%
\begin{tikzpicture}[node distance=1cm and 1.3cm]

% Shared column headers
\node[above, font=\bfseries] at (1.95, 3.3) {\textbf{Early Training}};
\node[above, font=\bfseries] at (7.15, 3.3) {\textbf{Mid Training}};
\node[above, font=\bfseries] at (12.35, 3.3) {\textbf{Late Training}};

% ============ STOCHASTIC MASKING (TOP ROW) ============
\node[left, font=\large\bfseries] at (-0.8, 1.8) {Global};

% Early training - stochastic
\node[unfrozen_node] (se1a) at (0, 2.8) {};
\node[unfrozen_node] (se1b) at (0, 2) {};
\node[unfrozen_node] (se1c) at (0, 1.2) {};
\node[unfrozen_node] (se1d) at (0, 0.4) {};

\node[unfrozen_node] (se2a) at (1.3, 2.4) {};
\node[frozen_node] (se2b) at (1.3, 1.6) {};
\node[unfrozen_node] (se2c) at (1.3, 0.8) {};

\node[unfrozen_node] (se3a) at (2.6, 2.1) {};
\node[unfrozen_node] (se3b) at (2.6, 1.1) {};

\node[unfrozen_node] (se4a) at (3.9, 2.1) {};
\node[unfrozen_node] (se4b) at (3.9, 1.1) {};

% Edges
\draw[active_edge] (se1a) -- (se2a);
\draw[blocked_edge] (se1a) -- (se2b);
\draw[active_edge] (se1b) -- (se2a);
\draw[blocked_edge] (se1b) -- (se2b);
\draw[active_edge] (se1b) -- (se2c);
\draw[active_edge] (se1c) -- (se2a);
\draw[blocked_edge] (se1c) -- (se2b);
\draw[active_edge] (se1c) -- (se2c);
\draw[frozen_edge] (se1d) -- (se2c);

\draw[active_edge] (se2a) -- (se3a);
\draw[active_edge] (se2a) -- (se3b);
\draw[active_edge] (se2b) -- (se3a);
\draw[active_edge] (se2b) -- (se3b);
\draw[active_edge] (se2c) -- (se3b);

\draw[active_edge] (se3a) -- (se4a);
\draw[active_edge] (se3a) -- (se4b);
\draw[active_edge] (se3b) -- (se4a);
\draw[active_edge] (se3b) -- (se4b);

% Mid training - stochastic
\node[unfrozen_node] (sm1a) at (5.2, 2.8) {};
\node[frozen_node] (sm1b) at (5.2, 2) {};
\node[unfrozen_node] (sm1c) at (5.2, 1.2) {};
\node[unfrozen_node] (sm1d) at (5.2, 0.4) {};

\node[frozen_node] (sm2a) at (6.5, 2.4) {};
\node[unfrozen_node] (sm2b) at (6.5, 1.6) {};
\node[frozen_node] (sm2c) at (6.5, 0.8) {};

\node[frozen_node] (sm3a) at (7.8, 2.1) {};
\node[unfrozen_node] (sm3b) at (7.8, 1.1) {};

\node[unfrozen_node] (sm4a) at (9.1, 2.1) {};
\node[unfrozen_node] (sm4b) at (9.1, 1.1) {};

% Edges
\draw[blocked_edge] (sm1a) -- (sm2a);
\draw[active_edge] (sm1a) -- (sm2b);
\draw[blocked_edge] (sm1b) -- (sm2a);
\draw[frozen_edge] (sm1b) -- (sm2b);
\draw[active_edge] (sm1b) -- (sm2c);
\draw[blocked_edge] (sm1c) -- (sm2a);
\draw[active_edge] (sm1c) -- (sm2b);
\draw[blocked_edge] (sm1c) -- (sm2c);
\draw[active_edge] (sm1d) -- (sm2b);
\draw[blocked_edge] (sm1d) -- (sm2c);

\draw[blocked_edge] (sm2a) -- (sm3a);
\draw[active_edge] (sm2a) -- (sm3b);
\draw[active_edge] (sm2b) -- (sm3a);
\draw[frozen_edge] (sm2b) -- (sm3b);
\draw[frozen_edge] (sm2c) -- (sm3a);
\draw[active_edge] (sm2c) -- (sm3b);

\draw[active_edge] (sm3a) -- (sm4a);
\draw[active_edge] (sm3a) -- (sm4b);
\draw[active_edge] (sm3b) -- (sm4a);
\draw[active_edge] (sm3b) -- (sm4b);

% Late training - stochastic
\node[frozen_node] (sl1a) at (10.4, 2.8) {};
\node[frozen_node] (sl1b) at (10.4, 2) {};
\node[unfrozen_node] (sl1c) at (10.4, 1.2) {};
\node[frozen_node] (sl1d) at (10.4, 0.4) {};

\node[frozen_node] (sl2a) at (11.7, 2.4) {};
\node[frozen_node] (sl2b) at (11.7, 1.6) {};
\node[unfrozen_node] (sl2c) at (11.7, 0.8) {};

\node[frozen_node] (sl3a) at (13, 2.1) {};
\node[unfrozen_node] (sl3b) at (13, 1.1) {};

\node[frozen_node] (sl4a) at (14.3, 2.1) {};
\node[unfrozen_node] (sl4b) at (14.3, 1.1) {};

% Edges
\draw[frozen_edge] (sl1a) -- (sl2a);
\draw[blocked_edge] (sl1a) -- (sl2b);
\draw[frozen_edge] (sl1b) -- (sl2a);
\draw[frozen_edge] (sl1b) -- (sl2b);
\draw[active_edge] (sl1b) -- (sl2c);
\draw[blocked_edge] (sl1c) -- (sl2a);
\draw[frozen_edge] (sl1c) -- (sl2b);
\draw[frozen_edge] (sl1c) -- (sl2c);
\draw[frozen_edge] (sl1d) -- (sl2a);
\draw[frozen_edge] (sl1d) -- (sl2b);
\draw[active_edge] (sl1d) -- (sl2c);

\draw[blocked_edge] (sl2a) -- (sl3a);
\draw[frozen_edge] (sl2a) -- (sl3b);
\draw[blocked_edge] (sl2b) -- (sl3a);
\draw[active_edge] (sl2b) -- (sl3b);
\draw[active_edge] (sl2c) -- (sl3b);
\draw[frozen_edge] (sl2c) -- (sl3a);

\draw[blocked_edge] (sl3a) -- (sl4a);
\draw[active_edge] (sl3b) -- (sl4a);
\draw[active_edge] (sl3b) -- (sl4b);
\draw[active_edge] (sl3a) -- (sl4b);

% ============ LAYER-WISE FORWARD MASKING (BOTTOM ROW) ============
\begin{scope}[yshift=2.5cm]

\node[left, font=\large\bfseries] at (-0.8, -4.2) {Layer-wise};

% Early training - forward
\node[frozen_node] (fe1a) at (0, -3.2) {};
\node[frozen_node] (fe1b) at (0, -4) {};
\node[frozen_node] (fe1c) at (0, -4.8) {};
\node[frozen_node] (fe1d) at (0, -5.6) {};

\node[unfrozen_node] (fe2a) at (1.3, -3.6) {};
\node[unfrozen_node] (fe2b) at (1.3, -4.4) {};
\node[unfrozen_node] (fe2c) at (1.3, -5.2) {};

\node[unfrozen_node] (fe3a) at (2.6, -3.9) {};
\node[unfrozen_node] (fe3b) at (2.6, -4.9) {};

\node[unfrozen_node] (fe4a) at (3.9, -3.9) {};
\node[unfrozen_node] (fe4b) at (3.9, -4.9) {};

% Edges
\draw[active_edge] (fe1a) -- (fe2a);
\draw[active_edge] (fe1a) -- (fe2b);
\draw[active_edge] (fe1b) -- (fe2a);
\draw[active_edge] (fe1b) -- (fe2b);
\draw[active_edge] (fe1b) -- (fe2c);
\draw[active_edge] (fe1c) -- (fe2b);
\draw[active_edge] (fe1c) -- (fe2c);
\draw[active_edge] (fe1d) -- (fe2b);
\draw[active_edge] (fe1d) -- (fe2c);

\draw[active_edge] (fe2a) -- (fe3a);
\draw[active_edge] (fe2a) -- (fe3b);
\draw[active_edge] (fe2b) -- (fe3a);
\draw[active_edge] (fe2b) -- (fe3b);
\draw[active_edge] (fe2c) -- (fe3b);

\draw[active_edge] (fe3a) -- (fe4a);
\draw[active_edge] (fe3a) -- (fe4b);
\draw[active_edge] (fe3b) -- (fe4a);
\draw[active_edge] (fe3b) -- (fe4b);

% Mid training - forward
\node[frozen_node] (fm1a) at (5.2, -3.2) {};
\node[frozen_node] (fm1b) at (5.2, -4) {};
\node[frozen_node] (fm1c) at (5.2, -4.8) {};
\node[frozen_node] (fm1d) at (5.2, -5.6) {};

\node[frozen_node] (fm2a) at (6.5, -3.6) {};
\node[frozen_node] (fm2b) at (6.5, -4.4) {};
\node[frozen_node] (fm2c) at (6.5, -5.2) {};

\node[unfrozen_node] (fm3a) at (7.8, -3.9) {};
\node[unfrozen_node] (fm3b) at (7.8, -4.9) {};

\node[unfrozen_node] (fm4a) at (9.1, -3.9) {};
\node[unfrozen_node] (fm4b) at (9.1, -4.9) {};

% Edges
\draw[frozen_edge] (fm1a) -- (fm2a);
\draw[frozen_edge] (fm1a) -- (fm2b);
\draw[frozen_edge] (fm1b) -- (fm2a);
\draw[frozen_edge] (fm1b) -- (fm2b);
\draw[frozen_edge] (fm1b) -- (fm2c);
\draw[frozen_edge] (fm1c) -- (fm2b);
\draw[frozen_edge] (fm1c) -- (fm2c);
\draw[frozen_edge] (fm1d) -- (fm2b);
\draw[frozen_edge] (fm1d) -- (fm2c);

\draw[frozen_edge] (fm2a) -- (fm3a);
\draw[frozen_edge] (fm2a) -- (fm3b);
\draw[frozen_edge] (fm2b) -- (fm3a);
\draw[frozen_edge] (fm2b) -- (fm3b);
\draw[frozen_edge] (fm2c) -- (fm3b);

\draw[active_edge] (fm3a) -- (fm4a);
\draw[active_edge] (fm3a) -- (fm4b);
\draw[active_edge] (fm3b) -- (fm4a);
\draw[active_edge] (fm3b) -- (fm4b);

% Late training - forward
\node[frozen_node] (fl1a) at (10.4, -3.2) {};
\node[frozen_node] (fl1b) at (10.4, -4) {};
\node[frozen_node] (fl1c) at (10.4, -4.8) {};
\node[frozen_node] (fl1d) at (10.4, -5.6) {};

\node[frozen_node] (fl2a) at (11.7, -3.6) {};
\node[frozen_node] (fl2b) at (11.7, -4.4) {};
\node[frozen_node] (fl2c) at (11.7, -5.2) {};

\node[frozen_node] (fl3a) at (13, -3.9) {};
\node[frozen_node] (fl3b) at (13, -4.9) {};

\node[unfrozen_node] (fl4a) at (14.3, -3.9) {};
\node[unfrozen_node] (fl4b) at (14.3, -4.9) {};

% Edges
\draw[frozen_edge] (fl1a) -- (fl2a);
\draw[frozen_edge] (fl1a) -- (fl2b);
\draw[frozen_edge] (fl1b) -- (fl2a);
\draw[frozen_edge] (fl1b) -- (fl2b);
\draw[frozen_edge] (fl1b) -- (fl2c);
\draw[frozen_edge] (fl1c) -- (fl2b);
\draw[frozen_edge] (fl1c) -- (fl2c);
\draw[frozen_edge] (fl1d) -- (fl2b);
\draw[frozen_edge] (fl1d) -- (fl2c);

\draw[frozen_edge] (fl2a) -- (fl3a);
\draw[frozen_edge] (fl2a) -- (fl3b);
\draw[frozen_edge] (fl2b) -- (fl3a);
\draw[frozen_edge] (fl2b) -- (fl3b);
\draw[frozen_edge] (fl2c) -- (fl3b);

\draw[frozen_edge] (fl3a) -- (fl4a);
\draw[frozen_edge] (fl3a) -- (fl4b);
\draw[frozen_edge] (fl3b) -- (fl4a);
\draw[frozen_edge] (fl3b) -- (fl4b);
\end{scope}

% ============ LEGEND ============
\begin{scope}[shift={(18, 0.5)}]  % Moved up 0.5 units
    % Legend box with rounded corners (no title needed)
    \node[draw, rectangle, rounded corners=5pt, thick, minimum width=3.2cm, minimum height=6.5cm, 
          fill=white, align=left] at (0, -0.5) {};
    
    % Unfrozen node
    \node[unfrozen_node] (leg_unfroz) at (-0.8, 2) {};
    \node[right, align=left] at (-0.3, 2) {Unfrozen\\node};
    
    % Frozen node
    \node[frozen_node] (leg_froz) at (-0.8, 0.8) {};
    \node[right, align=left] at (-0.3, 0.8) {Frozen\\node};
    
    % Active edge
    \draw[active_edge] (-1.2, -0.4) -- (-0.4, -0.4);
    \node[right, align=left] at (-0.3, -0.4) {Active\\gradient};
    
    % Frozen edge
    \draw[frozen_edge] (-1.2, -1.6) -- (-0.4, -1.6);
    \node[right, align=left] at (-0.3, -1.6) {Frozen\\gradient};
    
    % Blocked edge
    \draw[blocked_edge] (-1.2, -2.8) -- (-0.4, -2.8);
    \node[right, align=left] at (-0.3, -2.8) {Blocked\\gradient};
\end{scope}

\end{tikzpicture}
}

\caption{Comparison of masking strategies in StoMPP. \textbf{Top:} Global stochastic masking randomly freezes activations (nodes) and weights (edges) throughout training. \textbf{Bottom:} Layer-wise stochastic masking freezes entire layers sequentially from input to output. Blue indicates active gradient paths, red indicates frozen elements (sign function), and orange indicates edges blocked by frozen targets.}
\label{fig:masking_comparison}
\end{figure*}

\textbf{Masked Variables.}
For each scheduled layer $i$, StoMPP maintains either a weight mask $M_i^{W}$ with the same shape as weights $W_i$ for a weight layer, or an activation mask $M_i^{A}$ with the same shape as the layer pre-activations $z_i$ for an activation layer. We apply StoMPP sequentially to both weights and activations of scheduled layers (unless otherwise stated). For a weight tensor of size $n\times m$, $M_i^W$ has the same shape; for an activation vector of size $n$, $M_i^A$ is size $n$ (and analogously for convolutional tensors). Unless stated otherwise, masking is applied \emph{elementwise}.

\textbf{Freezing Schedule.}
Within a layer's transition, StoMPP targets an increasing frozen fraction $p(\tau)\in[0,1]$ over transition step $\tau=1,\ldots,T$. We use a cubic schedule ${p(\tau)=\left(\frac{\tau}{T}\right)^3}$
% \begin{equation}
%p(\tau)=\left(\frac{\tau}{T}\right)^3,
% \end{equation}
and recommend any monotonically increasing schedule from $0$ to $1$ ending in a fully binary layer ($p(T)=1$). Section~\ref{sec:hp_ablations} ablates the schedule shape.

\textbf{Soft Refresh.}
To prevent premature commitment to a particular frozen configuration, StoMPP \emph{soft-refreshes} the mask each step. For a tensor with $n$ entries, we resample only $k=\lfloor n/r\rfloor$ randomly chosen indices, redrawing those mask values from $\mathrm{Bernoulli}(p(\tau))$ while keeping all other indices unchanged; this preserves the target frozen fraction \emph{in expectation} and yields temporal stability (e.g., $r=100$ updates $\approx 1\%$ of entries per step). This differs from full resampling, where the entire mask changes each step, and from deterministic freezing (e.g., INQ-style progressive quantization), where frozen parameters never change. A slower refresh rate may allow model to adapt to a mostly stable frozen pattern before it is perturbed. Section~\ref{sec:hp_ablations} ablates $p(\tau)$ and $r$ and motivates our defaults. Algorithm~\ref{algo:layer_masking} summarizes the layer masking procedure, including the soft-refresh update and the resulting forward/backward computation.

% \begin{figure}[t]
% \begin{minipage}[t]{0.48\linewidth}
% \input{algorithms/layer_masking}
% \end{minipage}
% \hfill
% \begin{minipage}[t]{0.48\linewidth}
% \input{algorithms/stompp}
% \end{minipage}
% \end{figure}

% \subsection{Layerwise Scheduling to Prevent Gradient Blockades}
\subsection{Layerwise Scheduling for Stable Gradients}
\label{sec:layerwise-scheduling}
This section describes why globally freezing binary activations can obstruct learning in BNNs and how layerwise scheduling avoids this.

\begin{wrapfigure}{r}{0.5\linewidth}
\vspace{-1em}
\begin{minipage}{\linewidth}
\input{algorithms/stompp}
\vspace{-3em}
\end{minipage}
\end{wrapfigure}

\textbf{Gradient Blockade under Global Masking.}
In BNNs, freezing an activation replaces it with $a = \mathrm{sign}(z)$, whose derivative is zero almost everywhere. If activations are frozen at arbitrary depths (global masking), gradient signal to earlier layers can be severely attenuated or eliminated along many paths, creating a \textbf{gradient blockade}. This issue is specific to BNNs; BWNs do not have this problem. Their activations remain continuous (e.g., $\mathrm{clip}$), so gradient flow is preserved through activation layers no matter which weights are frozen. 
% In BNNs, freezing an activation uses $a=\mathrm{sign}(z)$, whose derivative is zero almost everywhere. If activations are frozen at arbitrary depths (global masking), gradient signal to earlier layers can be severely attenuated or eliminated along many paths, creating a \textbf{gradient blockade}. This issue is specific to BNNs: in binary weight networks where activations remain continuous (e.g., $\mathrm{clip}$), gradients can still propagate through activation layers. 

In particular, our unfrozen continuous activation proxy $\mathrm{clip}(z)$ has nonzero gradient for most unsaturated activations. Consider a binary activation layer with output $A = \mathrm{sign}(z)$ where $z$ is the pre-activation. Since $\frac{\partial \mathrm{sign}(z)}{\partial z} = 0$ almost everywhere, residual/skip connections and weight matrices may provide alternate routes, but scattered frozen activations can still substantially reduce usable gradient signal in practice. Fig.~\ref{fig:masking_comparison} illustrates this difference. With global masking (top), frozen activations may appear at arbitrary depths, and the first frozen activation on a path can eliminate gradient signal to earlier layers on that path. This motivates controlling \textit{where} binarization is applied over time.

\begin{wrapfigure}{r}{0.45\linewidth}
\vspace{-2.5em}
\begin{minipage}{\linewidth}
\input{algorithms/layer_masking}
\end{minipage}
\vspace{-1em}
\end{wrapfigure}

% Fig.~\ref{fig:masking_comparison} illustrates this difference. With global masking (top), frozen activations may appear at arbitrary depths, and the first frozen activation on a path can eliminate gradient signal to earlier layers on that path. This motivates controlling \emph{where} binarization is applied over time.

\textbf{Layerwise Schedule.}
StoMPP avoids blockades by binarizing layers sequentially from input to output. At any time, the network is partitioned into: (1) \textbf{Frozen prefix} ($1,\ldots,\ell-1$): fully binarized %($M_i^{W}=M_i^{A}=\mathbf{1}$)
($M_i=\mathbf{1}$)
; gradients to these layers are not required; (2) \textbf{Transition layer} ($\ell$): partially frozen, updated by SoftRefresh toward $p(\tau)$; and (3) \textbf{Unfrozen suffix} ($\ell+1,\ldots,N$): fully continuous %($M_i^{W}=M_i^{A}=\mathbf{0}$), 
($M_i=\mathbf{0}$)
providing a gradient path from the loss to layer $\ell$.
% \end{itemize}
This ensures the transitioning layer always receives a valid learning signal through its unfrozen entries (Eq.~\ref{eq:backward_ste}), while frozen layers are not exposed to gradient blockades. Fig.~\ref{fig:masking_comparison} (bottom) illustrates this: binarization advances as a contiguous input-to-output wave, so unfrozen downstream layers preserve a gradient path for the layer currently transitioning.

Algorithm~\ref{algo:stompp} gives the complete StoMPP procedure with layerwise scheduling, applying Algorithm~\ref{algo:layer_masking} sequentially to each layer. For simplicity, we allocate an equal number of epochs/steps to each layer's transition; exploring non-uniform schedules is left to future work.

\section{Experiments}
\label{sec:experiments}

\subsection{Experimental Setup}
\label{sec:exp_setup}

We evaluate StoMPP under a controlled protocol designed to isolate the effect of the binarization rule from confounding training-recipe choices. Unless stated otherwise, all methods share the same backbone, data preprocessing, augmentation, optimizer, batch size, training length, and evaluation procedure (See Appendix~\ref{sec:appendix_a}). The only method-specific difference is the binarization rule (vanilla STE, STE-free StoMPP, or StoMPP+STE).

\textbf{Architectures and Tasks.} Our main evaluation uses ResNet-18, ResNet-34, and ResNet-50 on CIFAR-10, CIFAR-100 \citep{cifar10_krizhevsky}, and ImageNet \citep{imagenet_russakovsky2015} to characterize depth scaling. To test whether the pattern generalizes beyond ResNet-style vision models, we additionally evaluate MobileNetV2 on CIFAR-100 (a non-ResNet CNN) and BERT-base fine-tuned on SST-2 \citep{SST2_Socher2013RecursiveDM, GLUE_Wang_Singh_Michael_Hill_Levy_Bowman_2018, Devlin_Chang_Lee_Toutanova_2019} (a transformer / non-vision setting). For BERT, we binarize feed-forward layer weights and activations while keeping attention projections in full precision. We also evaluate compatibility with binary-specific architectures via Bi-Real Net (Section~\ref{sec:bireal_modern}).

\textbf{Quantization Regimes.} We consider both BWN (binary weights, real-valued activations) and BNN (binary weights and activations). Following standard practice, the first and last layers and downsampling/projection layers are kept in full precision; all remaining layers follow the method-specific binarization rule. We report top-1 accuracy on the quantized network.

\textbf{Baseline.} Our primary baseline is BinaryConnect/BinaryNet-style STE \citep{BinaryConnect_Courbariaux_2016}: the forward pass binarizes via $\mathrm{sign}(\cdot)$, and the backward pass uses an identity surrogate. The STE baseline shares the architecture, recipe, and precision policy described above.

\textbf{Method Variants.} We evaluate two StoMPP variants: STE-free StoMPP, in which frozen entries receive zero gradient, and StoMPP+STE, in which frozen entries receive an STE surrogate. Both share the same layerwise progression. To isolate the contribution of the progression itself from any benefit conferred by the surrogate gradient, we conduct all ablations in Sections~\ref{sec:layerwise_ablation}-\ref{sec:lr_ablation} in the STE-free regime; the resulting characterization (ordering, schedule, refresh, dynamics) reflects the progression rather than its interaction with surrogate gradients.

\textbf{Training Recipe.}
For each dataset we start from a standard full-precision ResNet recipe (SGD, $lr=0.1$, $momentum=0.9$) and apply it uniformly to all binary methods, with two controlled deviations. \textit{(1) No weight decay.} In binary settings, $\ell_2$ regularization counteracts the desired concentration of weights near$\pm 1$ by pulling parameters toward zero. To avoid introducing a confound that different methods may tolerate differently, we disable weight decay for all binary runs (STE, STE-free StoMPP, and StoMPP+STE alike). \textit{(2) Constant learning rate in the main comparison.} Learning-rate schedules can interact strongly with progressive binarization, making it unclear whether improvements come from the binarization rule or from schedule tuning. We therefore use a constant learning rate for the main StoMPP-vs-STE comparison. Unless stated otherwise, StoMPP uses a layerwise progression over the binarized portion of the network (input to output), the cubic schedule $p(\tau) = (\tau / T)^3$, and refresh rate $r=100$. Section~\ref{sec:layerwise_ablation} ablates the progression order, and Section~\ref{sec:layerwise_ablation} ablates the schedule shape and refresh rate. See Appendix~\ref{sec:appendix_a} for per-dataset configurations.

\subsection{Main Results}
\label{sec:main_results}

\begin{table*}[b]
\centering
\caption{Top-1 test accuracy (\%) on quantized networks across architectures and datasets. Both STE-free StoMPP and StoMPP+STE substantially reduce the depth degradation observed under vanilla STE in the BNN setting, with the gap widening at depth; in the BWN setting, all three methods are comparatively close. ${\dagger}$ indicates collapsed, near-random, training.}
\label{tab:main_results}
\resizebox{\textwidth}{!}{%
\begin{tabular}{llccccccccccc}
\toprule
& & \multicolumn{3}{c}{\textbf{CIFAR-10}} & \multicolumn{3}{c}{\textbf{CIFAR-100}} & \multicolumn{3}{c}{\textbf{ImageNet}} & \textbf{SST-2} & \textbf{CIFAR-100} \\
\cmidrule(lr){3-5}\cmidrule(lr){6-8}\cmidrule(lr){9-11}\cmidrule(lr){12-12}\cmidrule(lr){13-13}
\textbf{Type} & \textbf{Method} & R18 & R34 & R50 & R18 & R34 & R50 & R18 & R34 & R50 & BERT & MobileNetV2 \\
 % & FP    & 91.8 & 92.0 & 88.0 & 71.1 & 71.8 & 68.3 & 66.3 & \textcolor{red}{TODO} & XXX & 91.5 & 70.7 \\
\midrule
\multicolumn{13}{c}{\cellcolor[HTML]{D0D0D0}\textbf{BWN}} \\
\midrule
\multirow{3}{*}{BWN}
  & STE          & 89.8 & {89.8} & 88.3 & 64.6 & 64.9 & 64.3 & \textbf{65.5} & --- & \textbf{67.8} & \textbf{84.0} & 65.3 \\
  & StoMPP       & {90.7} & 89.4 & \textbf{91.2} & 69.5 & 66.3 & 69.0 & 60.6 & --- & 67.3 & 83.7 & \textbf{67.7} \\
  & StoMPP + STE & \textbf{91.4} & \textbf{91.7} & 91.0 & \textbf{71.3} & \textbf{68.9} & \textbf{68.3} & 61.6 & --- & 65.9 & 83.9 & \textbf{67.7} \\
\midrule
\multicolumn{13}{c}{\cellcolor[HTML]{D0D0D0}\textbf{BNN}} \\
\midrule
\multirow{3}{*}{BNN}
  & STE          & 77.8 & 61.5 & 51.5 & 49.1 & 33.7 & 26.7 & {41.9} & 37.0 & 30.8 & 50.9$^{\dagger}$ & 40.5 \\
  & StoMPP       & 80.9 & 76.0 & 69.5 & 53.8 & 39.8 & 40.2 & 39.5 & 39.7 & 34.2 & 79.0 & 42.0 \\
  & StoMPP + STE & \textbf{86.1} & \textbf{82.1} & \textbf{78.6} & \textbf{58.0} & \textbf{47.8} & \textbf{46.5} & \textbf{45.5} & \textbf{48.1} & \textbf{48.5} & \textbf{82.8} & \textbf{48.0} \\
\bottomrule
\end{tabular}}
\end{table*}

\noindent\textbf{Binary Neural Networks (BNN).}
Table~\ref{tab:main_results} reports top-1 accuracy across architectures and datasets for vanilla STE, STE-free StoMPP, and StoMPP+STE. Vanilla STE drops sharply as the ResNets depth increases. On ImageNet, it falls from 41.9\% (R18) to 30.8\% (R50). StoMPP falls, but less sharply, from 39.5\% to 34.2\%. Only \textit{together} can StoMPP+STE overcome the depth scaling issue of binary neural networks, \textit{improving} from 45.5\% (R18) to 48.5\% (R50). 

We find a similar pattern on CIFAR-10/100. On CIFAR-10, vanilla STE it falls from 77.8\% (R18) to 51.5\% (R50), and on CIFAR-100 from 49.1\% (R18) to 26.7\% (R50). Both StoMPP variants flatten this drop. STE-free StoMPP improves over STE by +18.0/+13.5/+3.8 on CIFAR-10/100/ImageNet at R50, and StoMPP+STE further improves to +27.1/+19.8/+17.7 over STE on the same setting. The pattern holds at shallower depths (e.g., R18 CIFAR-100: STE 49.1, STE-free 53.8, StoMPP+STE 58.0) but the gap widens with depth. The pattern transfers beyond ResNets: on MobileNetV2 (CIFAR-100), STE-free StoMPP improves from 40.5 to 42.0 and StoMPP+STE reaches 48.0; on SST-2, STE collapses to 50.9 (near random for binary classification) while STE-free StoMPP reaches 79.0 and StoMPP+STE reaches 82.8.

Section~\ref{sec:layerwise_ablation}, analyzes this behavior through ordering/policy ablations and training dynamics.

\noindent\textbf{Binary Weight Networks (BWN).}
When activations remain continuous, all three methods are comparatively close, consistent with our claim that activation binarization is the dominant source of optimization difficulty. STE-free StoMPP matches or improves over STE in most configurations (CIFAR-100: 69.5/66.3/69.0 vs STE's 64.6/64.9/64.3 across R18/R34/R50), with smaller and less consistent gaps than in the BNN setting. StoMPP+STE shows similar behavior to STE-free StoMPP in BWN, supporting the interpretation that the surrogate-on-frozen variant primarily addresses BNN-specific blockades and offers little additional benefit when activations are already continuous.

\subsection{Ablation}
\label{sec:hp_ablations}

\textbf{Ordering Ablation: Layerwise Prevents Collapse.}
\label{sec:layerwise_ablation}
The progression's central design choice is the order in which layers are committed to binary. We compare three orderings under a fixed training recipe (Table~\ref{tab:stompp_masking_ablation}; Figure~\ref{fig:masking_comparison}). All ablations in this section use STE-free StoMPP to isolate the effect of ordering from any benefit conferred by surrogate gradients.

% Progressive freezing methods often apply a \emph{global} mask (e.g., INQ-style freezing) over all quantized layers. We find that this global masking is effective for BWNs but degrades or fails in BNNs. To isolate the role of progression order in the presence of binary activations, we compare three mask orderings under a fixed training recipe (Table~\ref{tab:stompp_masking_ablation}; Figure~\ref{fig:masking_comparison}).

\begin{wraptable}{r}{0.55\linewidth}
\vspace{-1em}
\centering
\small
\caption{\textbf{Masking Ordering and Policy for BWNs and BNNs.}
Reverse layerwise causes BNN failure with degradation at depth, isolating binary activations as the source of gradient blockades. Stochastic masking outperforms deterministic for BWN at depth. All models trained on CIFAR-100 for 200 epochs.
$\dagger$: catastrophic collapse; subscripts denote train/test gap.}
\label{tab:stompp_masking_ablation}
\resizebox{1\linewidth}{!}{
\begin{tabular}{llcccc}
\toprule
& & \multicolumn{2}{c}{\textbf{ResNet18}} & \multicolumn{2}{c}{\textbf{ResNet50}} \\
\cmidrule(lr){3-4}\cmidrule(lr){5-6}
\textbf{Schedule} & \textbf{Policy} & \textbf{Train} & \textbf{Test} & \textbf{Train} & \textbf{Test} \\
\midrule
\multicolumn{6}{c}{\cellcolor[HTML]{D0D0D0}\textbf{BNN}} \\
\midrule
Global          & Stochastic & 62.3 & 53.2$_{9.1}$  & 39.5 & 35.9$_{3.6}$ \\
Layerwise       & Stochastic & 83.3 & \textbf{53.8}$_{29.5}$ & 54.6 & \textbf{40.0}$_{14.6}$ \\
Reverse Layerwise & Stochastic & 28.4$^\dagger$ & 28.4$^\dagger_{0.0}$ & 9.6$^\dagger$ & 8.6$^\dagger_{1.0}$ \\
\midrule
\multicolumn{6}{c}{\cellcolor[HTML]{D0D0D0}\textbf{BWN}} \\
\midrule
\multirow{2}{*}{Global}
  & Deterministic & 99.5 & 70.2$_{29.3}$ & 96.9 & 65.9$_{31.0}$ \\
  & Stochastic    & 99.2 & \textbf{70.3}$_{28.9}$ & 96.1 & 67.5$_{28.6}$ \\
\multirow{2}{*}{Layerwise}
  & Deterministic & 99.9 & 69.6$_{30.3}$ & 99.8 & 65.5$_{34.3}$ \\
  & Stochastic    & 99.9 & 69.5$_{30.4}$ & 99.9 & \textbf{69.0}$_{30.9}$ \\
\multirow{2}{*}{Reverse Layerwise}
  & Deterministic & 99.0 & 68.9$_{30.1}$ & 99.7 & 68.4$_{31.3}$ \\
  & Stochastic    & 99.6 & 69.7$_{29.9}$ & 99.6 & 66.6$_{33.0}$ \\
\bottomrule
\vspace{2pt}
\end{tabular}
}
%\end{table}
\vspace{-2em}
\end{wraptable}

We evaluate three types of \textit{mask ordering}. (1) \textbf{Layerwise (input$\rightarrow$output):} progressively mask layers from the first quantized layer to the last. This is StoMPP’s default and is used in Section~\ref{sec:main_results}.
(2) \textbf{Global:} apply the same progression schedule to the entire quantized subnetwork at once (INQ-style).
(3) \textbf{Reverse layerwise (output$\rightarrow$input):} progressively mask layers from the last quantized layer to the first, intended to stress-test the effect of blocking gradient flow early in the network.
We use two masking policies: \textbf{(1) Stochastic:} freeze a fraction $p(t)$ at random and refresh a $\tfrac{1}{r}$ subset each step (default StoMPP); \textbf{(2) Deterministic (BWN only):} freeze the $p(t)\%$ of weights closest to $\pm1$ and refresh the frozen set each step. We do not apply (2) to BNNs since activations lack a “closeness to $\pm1$” analogue.

% \noindent\textit{Results and implication.}
In BNNs, the ordering is decisive: \textbf{layerwise} training yields the best performance, \textbf{global} is worse, and \textbf{reverse layerwise} causes catastrophic collapse (Table~\ref{tab:stompp_masking_ablation}). For example on CIFAR-100, reverse layerwise collapses to near-chance performance (R18: $28.4\%$, R50: $8.6\%$), while forward layerwise reaches $53.8\%$ (R18) and $40.0\%$ (R50). In contrast, BWNs are far less sensitive to ordering: both forward and reverse layerwise remain competitive, and deterministic masking can be effective at depth.
These results support the interpretation that \emph{binary activations} make training particularly sensitive to progression order: freezing later layers before earlier ones can severely restrict the effective learning signal reaching upstream quantized layers. The asymmetry sharpens with depth. The BNN gap between forward and reverse widens from R18 to R50, matching the prediction of Section 3.3 that gradient blockades introduced earlier in the network have more downstream impact in deeper models.

\begin{figure*}[tb]
    \centering
    \begin{overpic}[width=\textwidth]{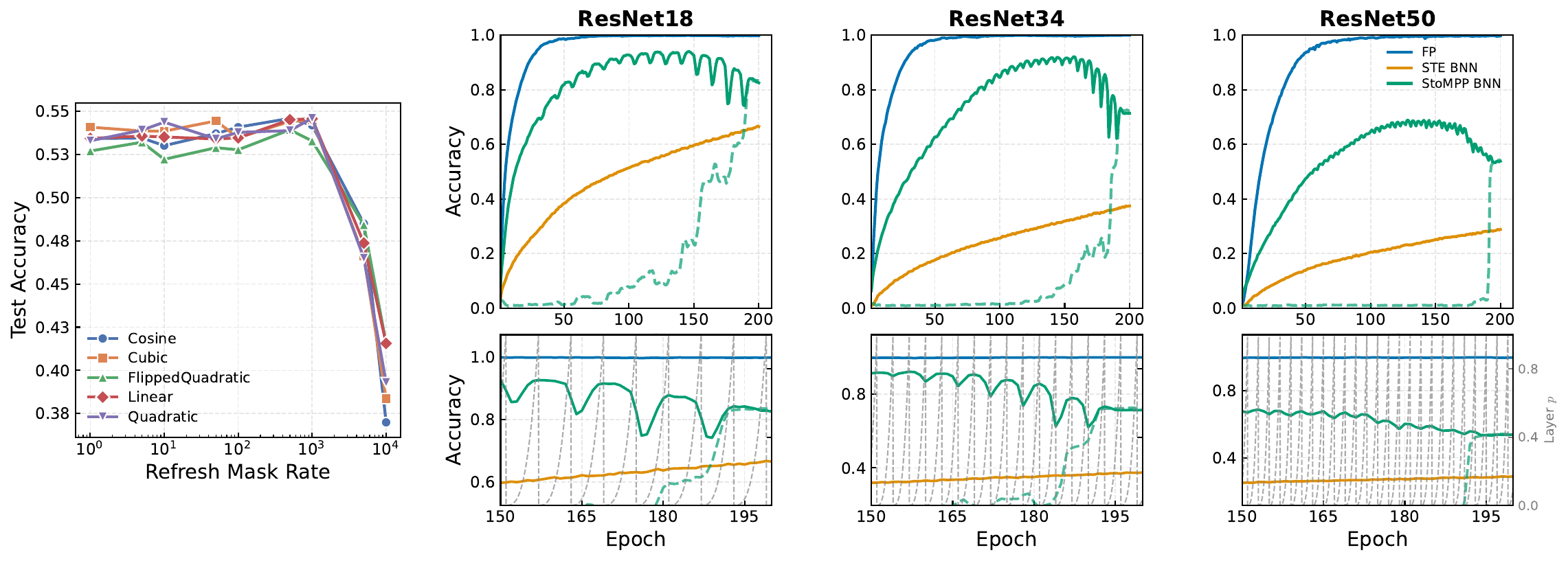}
        \put(15,-1.5){(a)}   % Centered below first subplot
        \put(40.7,-1.5){(b)}   % Centered below second subplot
        \put(64.9,-1.5){(c)}   % Centered below third subplot
        \put(88.8,-1.5){(d)}   % Centered below fourth subplot
    \end{overpic}
    \vspace{0.1em}
    \caption{
        % \textbf{(a)} Hyperparameter sweep for BNNs under the \emph{global} mask on CIFAR-100 with ResNet18. We vary the freezing schedule $p(t)$ and refresh rate $r$ for StoMPP and report Top-1 test accuracy (\%).
        % \textbf{(b--d)} Accuracy trajectories of CIFAR-100 on ResNets, trained with STE and StoMPP under the same training recipe. StoMPP exhibits a sawtooth pattern corresponding to progressive freezing, while STE improves more smoothly over training. The dashed line represents the fully quantized StoMPP network accuracy. \textcolor{red}{TODO fix this prose to fit with the more stable layer-wise version, because that version is much more stable, describe insets}
        \textbf{(a)} Hyperparameter sweep for BNNs under the \emph{layerwise} mask on CIFAR-100 with ResNet18. We vary the freezing schedule $p(t)$ and refresh rate $r$ for StoMPP and report Top-1 test accuracy (\%).
        \textbf{(b--d)} Accuracy trajectories of CIFAR-100 on ResNets, trained with STE and StoMPP under the same training recipe. StoMPP exhibits a sawtooth pattern corresponding to progressive freezing, while STE improves more smoothly over training. The dashed line represents the fully quantized StoMPP network accuracy. The gray line represents the percentage of the current layer frozen throughout layerwise freezing.
    }
    \label{fig:hp_sweep_and_curves}
    \vspace{-2em}
\end{figure*}
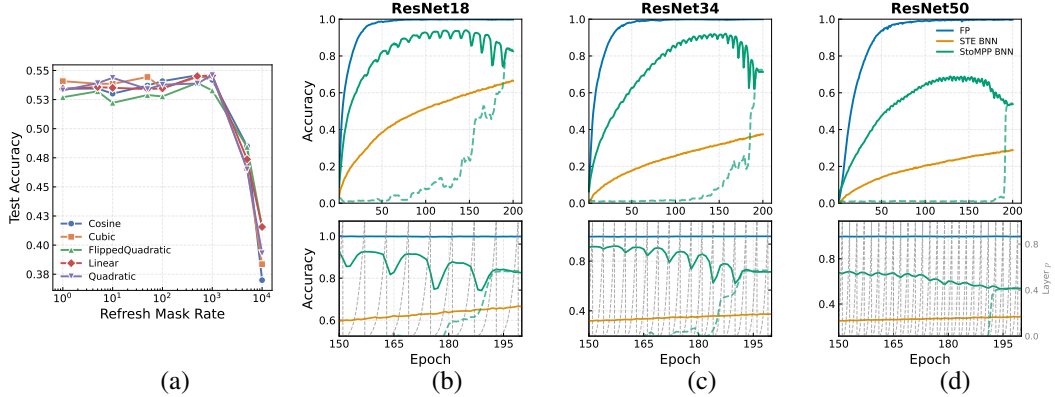

\textbf{Hyperparameter Ablation: Schedule \& Refresh.}
StoMPP's progression is governed by two hyperparameters: a freezing schedule $p(t) \in [0, 1]$ that specifies the frozen fraction at training step $t$, and a refresh rate $r$ that controls how often the frozen set is resampled (approximately $1/r$ of masked indices per step). 

% \textit{Default Selection Protocol.} 
All ablations in this section are conducted on CIFAR-100 ResNet-18, and the resulting defaults (cubic schedule, $r = 100$) are then applied without retuning across all other architectures and datasets. This avoids per-setting hyperparameter tuning that could confound the main comparison.

% \textbf{Schedules.} 
Let $T$ denote the total number of progression steps, and $t \in \{0, \ldots, T\}$ the current step. We compare five monotone schedules: $\textit{Cosine}\;p(t) = \tfrac{1}{2} - \tfrac{1}{2}\cos(\pi t / T)$, $\textit{Linear}\;p(t) = t/T$, $\textit{Quadratic}\;p(t) = (t/T)^2$, $\textit{Cubic}\;p(t) = (t/T)^3$, and $\textit{Flipped quadratic}\;p(t) = 2(t/T) - (t/T)^2$.

% Older version referencing global results
% \textbf{Results.} Figure~\ref{fig:hp_sweep_and_curves} (a) sweeps schedule shape and refresh rate under both layerwise and global progression. Two findings emerge. First, performance depends strongly on refresh rate: moderate values ($r \in [10^2, 10^3]$) work well across schedules, while very large values ($r = 10^4$) cause training to stall or collapse, indicating that insufficient refreshing prevents the model from adapting to the frozen pattern. Second, layerwise progression is substantially more robust to schedule and refresh choice than global progression. Under global progression, schedule shape matters: the cubic schedule is the best performer over a broad range, while linear and flipped-quadratic schedules degrade earlier as $r$ increases. Under layerwise progression, all five schedules perform similarly across most of the sweep, with degradation only at very high refresh rates.

% This robustness is consistent with the layerwise schedule's role: by ensuring an unfrozen suffix at all times, it preserves a gradient path regardless of the exact rate at which entries within the transition layer are committed. We use the cubic schedule and $r=100$ as defaults throughout the paper, which lie in the stable high-performing region of both sweeps.

% \textcolor{blue}{
% \textbf{Results.} 
Figure~\ref{fig:hp_sweep_and_curves} (a) sweeps schedule shape and refresh rate under layerwise masking progression. Across many monotonically increasing schedules and reasonable refresh rates (<$10^3$) StoMPP remains stable.
% Two findings emerge. First, performance depends strongly on refresh rate: moderate values ($r \in [10^2, 10^3]$) work well across schedules, while very large values ($r = 10^4$) cause training to stall or collapse, indicating that insufficient refreshing prevents the model from adapting to the frozen pattern. Second, layerwise progression is substantially more robust to schedule and refresh choice than global progression. Under global progression, schedule shape matters: the cubic schedule is the best performer over a broad range, while linear and flipped-quadratic schedules degrade earlier as $r$ increases. Under layerwise progression, all five schedules perform similarly across most of the sweep, with degradation only at very high refresh rates.
This robustness is consistent with the layerwise schedule's role: by ensuring an unfrozen suffix at all times, it preserves a gradient path regardless of the exact rate at which entries within the transition layer are committed. We use the cubic schedule and $r=100$ as defaults throughout the paper, which lie in the stable high-performing region of this sweep. These defaults are also most effective for the global masking explored in Section~\ref{sec:layerwise_ablation}, where these hyperparameters are more sensitive. For most details on these hyperparameters in a global masking setting, see Appendix~\ref{sec:app_hp_forward_masking}.

\textbf{Dynamics Ablation: Sawtooth Training Dynamics.}
\label{sec:training_dynamics}
Figure~\ref{fig:hp_sweep_and_curves} illustrates that STE-free StoMPP and STE exhibit qualitatively different optimization dynamics on CIFAR-100 across network depths. Under STE, test accuracy typically increases smoothly over training, consistent with a fixed binarized forward pass throughout optimization. In contrast, StoMPP produces a characteristic \emph{sawtooth} trajectory: when a new layer enters the freezing phase, accuracy drops, followed by a recovery period as the remaining unfrozen parameters adapt. This pattern repeats as StoMPP progresses through the network until all scheduled layers are binarized.
These dynamics are also reflected in the hybrid ablation (Appendix~\ref{sec:hybrids}). The variant that applies StoMPP to weights and STE to activations exhibits the same sawtooth behavior, whereas the variant that applies StoMPP only to activations does not, consistent with StoMPP's primary effect operating through the progression of weights. StoMPP+STE produces qualitatively similar sawtooth dynamics, since the layerwise progression dominates the trajectory in both variants; see Appendix~\ref{sec:appendix_b} for curves.
% These dynamics are also reflected in the hybrid ablation (Section~\ref{sec:hybrids}): the reverse hybrid (StoMPP weights, STE activations) exhibits the same sawtooth behavior, whereas the hybrid that applies StoMPP only to activations does not, consistent with StoMPP’s primary effect operating through the progression/freezing of weights. We report additional training curves (including BWN variants) and corresponding training-accuracy trajectories in Appendix~\ref{sec:appendix_b}.

\textbf{Epoch Ablation: Accuracy and Training Epochs.}
\label{sec:epoch_ablation}
Longer schedules are common for BNNs, so we sweep training epochs and compare STE vs. StoMPP on ResNet18/50, reporting train and test accuracy to separate optimization from generalization. Figure~\ref{fig:epoch_ablation} shows both methods benefit from more epochs, while the full-precision reference converges much faster. StoMPP achieves higher training accuracy early, indicating faster optimization in the low-epoch regime. With sufficient training, STE partially closes the gap on ResNet18 (e.g., by 500 epochs), but on deeper ResNet50 it improves more slowly and remains lower throughout, consistent with stronger depth-induced optimization difficulty. Overall, StoMPP converges in fewer epochs and scales better with depth under the same recipe.

\textbf{Learning-Rate Ablation: Sensitivity Analysis.}
\label{sec:lr_ablation}
We ablate the learning rate with all other settings fixed, sweeping $\mathrm{lr}\in{10^{-3},10^{-2},10^{-1}}$ to cover typical BNN/QAT practice~\citep{QNN_Hubara_2016}. We exclude larger rates since they often require extra stabilization (e.g., warmup) that would confound a controlled comparison~\citep{Oscillations_2022}. Table~\ref{tab:lr_sweep} shows StoMPP outperforms STE across learning rates and prefers a slightly smaller optimum, suggesting StoMPP may benefit from tailored recipes.

\begin{figure*}[t]
\centering
\begin{minipage}[t]{0.55\linewidth}
    \vspace{0pt}
    \centering
    \includegraphics[width=\linewidth]{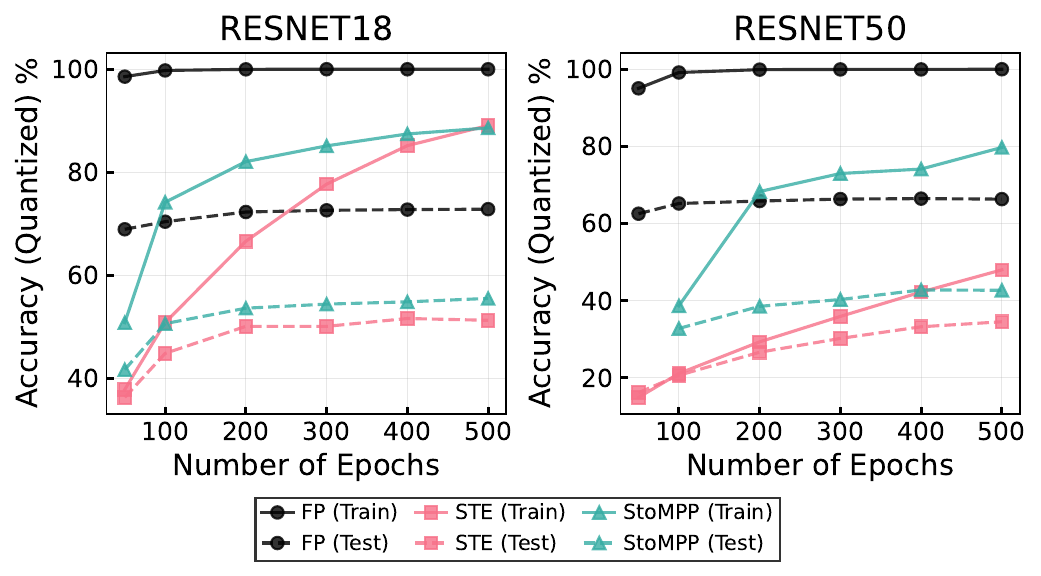}
    \captionof{figure}{Sweep of epochs under training scheme; solid lines show 
    train accuracy and dashed lines show test accuracy. Note 50 epochs is not 
    included for StoMPP ResNet50 as there are not enough epochs to apply at 
    least one epoch per layer of layerwise masking.}
    \label{fig:epoch_ablation}
\end{minipage}
\hfill
\begin{minipage}[t]{0.43\linewidth}
    \vspace{0pt}
    \centering
    % %\begin{table}[t]

% \centering
% %\caption
% \footnotesize
% \captionof{table}{\textbf{Learning Rate Sensitivity on CIFAR-100 ResNet-18 BNN.} 
% StoMPP is more stable at low learning rates (33.2\% at lr=0.001 vs. STE’s 10.4\%) and reaches a higher peak at lr=0.05 (54.2\% vs. 48.9\%). All models are trained for 200 epochs with a constant learning rate; results report top-1 test accuracy.}
% \label{tab:lr_sweep}
% \small
% \begin{tabular}{@{}ccccc@{}}
% \toprule
% \multicolumn{1}{l}{Learning Rate} & \multicolumn{1}{l}{0.001} & \multicolumn{1}{l}{0.01} & \multicolumn{1}{l}{0.05} & \multicolumn{1}{l}{0.1} \\ \midrule
% STE                               & 10.4                      & 30.9                     & 46.8                     & \textbf{48.9}           \\
% StoMPP                            & 33.2                      & 51.5                     & \textbf{54.2}            & 53.8                    \\ \bottomrule
% \end{tabular}
% %\end{table}

\captionof{table}{\textbf{Learning Rate Sensitivity on CIFAR-100 ResNet-18 BNN.} 
%\textcolor{red}{Can this be cut down, put elsewhere to make figure work?: StoMPP is more stable at low learning rates (33.2\% at lr=0.001 vs.\ STE's 10.4\%) and reaches a higher peak at lr=0.05 (54.2\% vs.\ 48.9\%).} 
All models trained for 
200 epochs with a constant learning rate; results report top-1 test accuracy.}
\label{tab:lr_sweep}
\small
\resizebox{\columnwidth}{!}{%
\begin{tabular}{@{}ccccc@{}}
\toprule
Learning Rate & 0.001 & 0.01 & 0.05 & 0.1 \\
\midrule
STE    & 10.4 & 30.9 & 46.8 & \textbf{48.9} \\
StoMPP & 33.2 & 51.5 & \textbf{54.2} & 53.8 \\
\bottomrule
\end{tabular}
}

    \vspace{1.5em}

\captionof{table}{Top-1 accuracy (\%) on CIFAR-100 for BiReal18/34.
%\textcolor{red}{Can this sentence be chopped (put elsewhere): Bi-Real Net primarily reports ImageNet settings; for CIFAR we follow the standard modified ResNet stem~\cite{ResNets_He_2015, BiReal_Liu_2018}.}
}
\label{tab:bireal}
\small
\resizebox{\columnwidth}{!}{%
\begin{tabular}{lcccc}
\toprule
Training & \multicolumn{2}{c}{BiReal18} & \multicolumn{2}{c}{BiReal34} \\
\cmidrule(lr){2-3}\cmidrule(lr){4-5}
         & StoMPP & STE & StoMPP & STE \\
\midrule
Scratch    & 56.6 & 56.0 & 59.1 & 55.8 \\
Pretrained & 59.4 & 63.5 & 62.3 & 60.6 \\
\bottomrule
\end{tabular}
}
\end{minipage}
\end{figure*}

% \begin{figure}[t]
%     \centering
%     \includegraphics[width=0.75\linewidth]{figures/epoch_ablation.pdf}
%     \caption{Sweep of epochs under training scheme; solid lines show train accuracy and dashed lines show test accuracy. Note 50 epochs is not included for StoMPP ResNet50 as there are not enough epochs to apply at least one epoch per layer of layerwise masking.}
%     \label{fig:epoch_ablation}
% \end{figure}

% \input{tables/lr_ablation}
% \input{tables/birealnet}

\textbf{Compatibility with Binary Architectures.}
\label{sec:bireal_modern}
We test whether StoMPP transfers to a binary-specific architecture by combining it with Bi-Real Net \citep{BiReal_Liu_2018}. Holding the architecture fixed, we vary only the training rule (STE vs.\ STE-free StoMPP) under matched compute (l$r =0.06$, $300$ epochs). See Appendix~\ref{sec:appendix_a} for more training details. Table~\ref{tab:bireal} reports CIFAR-100 results for BiReal-18 and BiReal-34, both from scratch and finetuned from a pretrained continuous network. From scratch, StoMPP matches or improves over STE on both architectures. When finetuning, the ranking flips on BiReal-18 (StoMPP 59.4 vs.\ STE 63.5) while StoMPP still improves on BiReal-34. Bi-Real's added skip connections also moderate StoMPP's sawtooth dynamics (Appendix \ref{sec:appendix_b}), since the residual paths supply gradient signal as upstream layers binarize. Overall, StoMPP transfers to binary-specific architectures without modification, though the interaction with finetuning suggests training rule and initialization are not independent.

% \textcolor{red}{old}
% We evaluate whether StoMPP’s gains persist on top of a binary-specific architecture (Bi-Real Net).

% \textcolor{red}{NOTE: I think this result (OvSW on teh UNFROZEN part of the mask, is way too confusing and unhelpful to include in this revision), so dropping this might help. and how it interact with STE refinements (e.g. OvSW~\cite{OvSW_Xiang}).}

% We train BiReal18/34 from scratch and finetune with an identical recipe (same optimizer/augmentation; $\mathrm{lr}=0.06$, 300 epochs). Holding the architecture fixed, we vary only the training rule (StoMPP and improved STE; Section~\ref{sec:exp_setup}) under matched compute. Table~\ref{tab:bireal} shows StoMPP matches or improves over STE, indicating it transfers to binary-specific architectures without additional architectural changes.

% \textcolor{red}{TODO ADD BERT TEXT HERE (or move BERT to main results, which I believe makes more sense.)}

% \textcolor{red}{TODO MAYBE REMOVE THIS, IT IS CONFUSING}
% We combine StoMPP with OvSW by applying OvSW only to the unfrozen parameters. On CIFAR-10, OvSW improves the STE baseline to 91.3\%, but drops to 9.85\% when paired with StoMPP. This suggests OvSW’s gains are not additive under a naive composition, motivating progressive freezing compatible refinements.

\section{Conclusion}
We introduced StoMPP, a layerwise progressive freezing procedure for training binary networks. The progression is independent of the choice of frozen-entry gradient: setting it to zero gives an STE-free training procedure, while applying STE only to frozen entries (StoMPP+STE) further improves accuracy. Our central empirical finding is that progression order is decisive when activations are binarized: forward layerwise progression succeeds on deep BNNs, reverse layerwise collapses to near-chance, and BWNs show no comparable sensitivity. Under matched training recipes, both StoMPP variants improve over vanilla STE with gains that grow with depth, and the pattern transfers from ResNets to MobileNetV2 and to BERT fine-tuning.

\section{Limitations and Discussion}
\label{sec:limitations}
Our evidence for activation-induced gradient blockades as the operative mechanism is consistent with our experiments but not isolated by them. The progression introduces hyperparameters absent from vanilla STE (schedule shape, refresh rate, per-layer step allocation), though layerwise scheduling is substantially more robust to these than global scheduling. We use a matched recipe to isolate the binarization rule, so absolute accuracies are below SOTA reported by methods that combine surrogate gradients with distillation, learning-rate schedules, or architectural changes. Future work includes designing a direct intervention to test the blockade mechanism, integrating StoMPP with competitive recipes, and broader transformer evaluation.

\bibliographystyle{abbrvnat}
\bibliography{stompp}

@article{Bai_Wang_Liberty_2019a, title={ProxQuant: Quantized Neural Networks via Proximal Operators}, url={http://arxiv.org/abs/1810.00861}, DOI={10.48550/arXiv.1810.00861}, abstractNote={To make deep neural networks feasible in resource-constrained environments (such as mobile devices), it is beneficial to quantize models by using low-precision weights. One common technique for quantizing neural networks is the straight-through gradient method, which enables back-propagation through the quantization mapping. Despite its empirical success, little is understood about why the straight-through gradient method works. Building upon a novel observation that the straight-through gradient method is in fact identical to the well-known Nesterov’s dual-averaging algorithm on a quantization constrained optimization problem, we propose a more principled alternative approach, called ProxQuant, that formulates quantized network training as a regularized learning problem instead and optimizes it via the prox-gradient method. ProxQuant does back-propagation on the underlying full-precision vector and applies an efficient prox-operator in between stochastic gradient steps to encourage quantizedness. For quantizing ResNets and LSTMs, ProxQuant outperforms state-of-the-art results on binary quantization and is on par with state-of-the-art on multi-bit quantization. For binary quantization, our analysis shows both theoretically and experimentally that ProxQuant is more stable than the straight-through gradient method (i.e. BinaryConnect), challenging the indispensability of the straight-through gradient method and providing a powerful alternative.}, note={arXiv:1810.00861}, number={arXiv:1810.00861}, publisher={arXiv}, author={Bai, Yu and Wang, Yu-Xiang and Liberty, Edo}, year={2019}, month=mar }

@article{AlphaBlend_Liu_Mattina_2019, title={Learning low-precision neural networks without Straight-Through Estimator(STE)}, url={http://arxiv.org/abs/1903.01061}, DOI={10.48550/arXiv.1903.01061}, abstractNote={The Straight-Through Estimator (STE) is widely used for back-propagating gradients through the quantization function, but the STE technique lacks a complete theoretical understanding. We propose an alternative methodology called alpha-blending (AB), which quantizes neural networks to low-precision using stochastic gradient descent (SGD). Our method (AB) avoids STE approximation by replacing the quantized weight in the loss function by an affine combination of the quantized weight w_q and the corresponding full-precision weight w with non-trainable scalar coefficient $α$ and $1-α$. During training, $α$ is gradually increased from 0 to 1; the gradient updates to the weights are through the full-precision term, $(1-α)w$, of the affine combination; the model is converted from full-precision to low-precision progressively. To evaluate the method, a 1-bit BinaryNet on CIFAR10 dataset and 8-bits, 4-bits MobileNet v1, ResNet_50 v1/2 on ImageNet dataset are trained using the alpha-blending approach, and the evaluation indicates that AB improves top-1 accuracy by 0.9%, 0.82% and 2.93% respectively compared to the results of STE based quantization.}, note={arXiv:1903.01061}, number={arXiv:1903.01061}, publisher={arXiv}, author={Liu, Zhi-Gang and Mattina, Matthew}, year={2019}, month=may }

@article{cifar10_krizhevsky,
author = {Krizhevsky, Alex},
year = {2012},
month = {05},
pages = {},
title = {Learning Multiple Layers of Features from Tiny Images},
journal = {University of Toronto}
}

@article{imagenet_russakovsky2015,
  title={ImageNet Large Scale Visual Recognition Challenge},
  author={Russakovsky, Olga and Deng, Jia and Su, Hao and Krause, Jonathan and Satheesh, Sanjeev and Ma, Sean and Huang, Zhiheng and Karpathy, Andrej and Khosla, Aditya and Bernstein, Michael and Berg, Alexander C and Fei-Fei, Li},
  journal={arXiv preprint arXiv:1409.0575},
  year={2014}
}

@article{ste_bengio_2013, title={Estimating or Propagating Gradients Through Stochastic Neurons for Conditional Computation}, url={http://arxiv.org/abs/1308.3432}, DOI={10.48550/arXiv.1308.3432}, abstractNote={Stochastic neurons and hard non-linearities can be useful for a number of reasons in deep learning models, but in many cases they pose a challenging problem: how to estimate the gradient of a loss function with respect to the input of such stochastic or non-smooth neurons? I.e., can we “back-propagate” through these stochastic neurons? We examine this question, existing approaches, and compare four families of solutions, applicable in different settings. One of them is the minimum variance unbiased gradient estimator for stochatic binary neurons (a special case of the REINFORCE algorithm). A second approach, introduced here, decomposes the operation of a binary stochastic neuron into a stochastic binary part and a smooth differentiable part, which approximates the expected effect of the pure stochatic binary neuron to first order. A third approach involves the injection of additive or multiplicative noise in a computational graph that is otherwise differentiable. A fourth approach heuristically copies the gradient with respect to the stochastic output directly as an estimator of the gradient with respect to the sigmoid argument (we call this the straight-through estimator). To explore a context where these estimators are useful, we consider a small-scale version of {em conditional computation}, where sparse stochastic units form a distributed representation of gaters that can turn off in combinatorially many ways large chunks of the computation performed in the rest of the neural network. In this case, it is important that the gating units produce an actual 0 most of the time. The resulting sparsity can be potentially be exploited to greatly reduce the computational cost of large deep networks for which conditional computation would be useful.}, note={arXiv:1308.3432}, number={arXiv:1308.3432}, publisher={arXiv}, author={Bengio, Yoshua and Léonard, Nicholas and Courville, Aaron}, year={2013}, month=aug }

@article{BinaryConnect_Courbariaux_2016, title={Binarized Neural Networks: Training Deep Neural Networks with Weights and Activations Constrained to +1 or -1}, url={http://arxiv.org/abs/1602.02830}, DOI={10.48550/arXiv.1602.02830}, abstractNote={We introduce a method to train Binarized Neural Networks (BNNs) - neural networks with binary weights and activations at run-time. At training-time the binary weights and activations are used for computing the parameters gradients. During the forward pass, BNNs drastically reduce memory size and accesses, and replace most arithmetic operations with bit-wise operations, which is expected to substantially improve power-efficiency. To validate the effectiveness of BNNs we conduct two sets of experiments on the Torch7 and Theano frameworks. On both, BNNs achieved nearly state-of-the-art results over the MNIST, CIFAR-10 and SVHN datasets. Last but not least, we wrote a binary matrix multiplication GPU kernel with which it is possible to run our MNIST BNN 7 times faster than with an unoptimized GPU kernel, without suffering any loss in classification accuracy. The code for training and running our BNNs is available on-line.}, note={arXiv:1602.02830}, number={arXiv:1602.02830}, publisher={arXiv}, author={Courbariaux, Matthieu and Hubara, Itay and Soudry, Daniel and El-Yaniv, Ran and Bengio, Yoshua}, year={2016}, month=mar }

@article{Soft_then_hard_Guo_2024, title={Soft then Hard: Rethinking the Quantization in Neural Image Compression}, url={http://arxiv.org/abs/2104.05168}, DOI={10.48550/arXiv.2104.05168}, abstractNote={Quantization is one of the core components in lossy image compression. For neural image compression, end-to-end optimization requires differentiable approximations of quantization, which can generally be grouped into three categories: additive uniform noise, straight-through estimator and soft-to-hard annealing. Training with additive uniform noise approximates the quantization error variationally but suffers from the train-test mismatch. The other two methods do not encounter this mismatch but, as shown in this paper, hurt the rate-distortion performance since the latent representation ability is weakened. We thus propose a novel soft-then-hard quantization strategy for neural image compression that first learns an expressive latent space softly, then closes the train-test mismatch with hard quantization. In addition, beyond the fixed integer quantization, we apply scaled additive uniform noise to adaptively control the quantization granularity by deriving a new variational upper bound on actual rate. Experiments demonstrate that our proposed methods are easy to adopt, stable to train, and highly effective especially on complex compression models.}, note={arXiv:2104.05168}, number={arXiv:2104.05168}, publisher={arXiv}, author={Guo, Zongyu and Zhang, Zhizheng and Feng, Runsen and Chen, Zhibo}, year={2024}, month=mar }

@article{Self_Binarizing_Lahoud_2019, title={Self-Binarizing Networks}, url={http://arxiv.org/abs/1902.00730}, DOI={10.48550/arXiv.1902.00730}, abstractNote={We present a method to train self-binarizing neural networks, that is, networks that evolve their weights and activations during training to become binary. To obtain similar binary networks, existing methods rely on the sign activation function. This function, however, has no gradients for non-zero values, which makes standard backpropagation impossible. To circumvent the difficulty of training a network relying on the sign activation function, these methods alternate between floating-point and binary representations of the network during training, which is sub-optimal and inefficient. We approach the binarization task by training on a unique representation involving a smooth activation function, which is iteratively sharpened during training until it becomes a binary representation equivalent to the sign activation function. Additionally, we introduce a new technique to perform binary batch normalization that simplifies the conventional batch normalization by transforming it into a simple comparison operation. This is unlike existing methods, which are forced to the retain the conventional floating-point-based batch normalization. Our binary networks, apart from displaying advantages of lower memory and computation as compared to conventional floating-point and binary networks, also show higher classification accuracy than existing state-of-the-art methods on multiple benchmark datasets.}, note={arXiv:1902.00730}, number={arXiv:1902.00730}, publisher={arXiv}, author={Lahoud, Fayez and Achanta, Radhakrishna and Márquez-Neila, Pablo and Süsstrunk, Sabine}, year={2019}, month=feb }

@article{ReActNet_Liu_2020, title={ReActNet: Towards Precise Binary Neural Network with Generalized Activation Functions}, url={http://arxiv.org/abs/2003.03488}, DOI={10.48550/arXiv.2003.03488}, abstractNote={In this paper, we propose several ideas for enhancing a binary network to close its accuracy gap from real-valued networks without incurring any additional computational cost. We first construct a baseline network by modifying and binarizing a compact real-valued network with parameter-free shortcuts, bypassing all the intermediate convolutional layers including the downsampling layers. This baseline network strikes a good trade-off between accuracy and efficiency, achieving superior performance than most of existing binary networks at approximately half of the computational cost. Through extensive experiments and analysis, we observed that the performance of binary networks is sensitive to activation distribution variations. Based on this important observation, we propose to generalize the traditional Sign and PReLU functions, denoted as RSign and RPReLU for the respective generalized functions, to enable explicit learning of the distribution reshape and shift at near-zero extra cost. Lastly, we adopt a distributional loss to further enforce the binary network to learn similar output distributions as those of a real-valued network. We show that after incorporating all these ideas, the proposed ReActNet outperforms all the state-of-the-arts by a large margin. Specifically, it outperforms Real-to-Binary Net and MeliusNet29 by 4.0% and 3.6% respectively for the top-1 accuracy and also reduces the gap to its real-valued counterpart to within 3.0% top-1 accuracy on ImageNet dataset. Code and models are available at: https://github.com/liuzechun/ReActNet.}, note={arXiv:2003.03488}, number={arXiv:2003.03488}, publisher={arXiv}, author={Liu, Zechun and Shen, Zhiqiang and Savvides, Marios and Cheng, Kwang-Ting}, year={2020}, month=july }

@misc{BNN_Survey_2020, title={Binary Neural Networks: A Survey}, url={https://arxiv.org/abs/2004.03333v1}, DOI={10.1016/j.patcog.2020.107281}, abstractNote={The binary neural network, largely saving the storage and computation, serves as a promising technique for deploying deep models on resource-limited devices. However, the binarization inevitably causes severe information loss, and even worse, its discontinuity brings difficulty to the optimization of the deep network. To address these issues, a variety of algorithms have been proposed, and achieved satisfying progress in recent years. In this paper, we present a comprehensive survey of these algorithms, mainly categorized into the native solutions directly conducting binarization, and the optimized ones using techniques like minimizing the quantization error, improving the network loss function, and reducing the gradient error. We also investigate other practical aspects of binary neural networks such as the hardware-friendly design and the training tricks. Then, we give the evaluation and discussions on different tasks, including image classification, object detection and semantic segmentation. Finally, the challenges that may be faced in future research are prospected.}, journal={arXiv.org}, author={Qin, Haotong and Gong, Ruihao and Liu, Xianglong and Bai, Xiao and Song, Jingkuan and Sebe, Nicu}, year={2020}, month=mar, language={en} }

@article{XNOR_Net_Rastegari_2016, title={XNOR-Net: ImageNet Classification Using Binary Convolutional Neural Networks}, url={http://arxiv.org/abs/1603.05279}, DOI={10.48550/arXiv.1603.05279}, abstractNote={We propose two efficient approximations to standard convolutional neural networks: Binary-Weight-Networks and XNOR-Networks. In Binary-Weight-Networks, the filters are approximated with binary values resulting in 32x memory saving. In XNOR-Networks, both the filters and the input to convolutional layers are binary. XNOR-Networks approximate convolutions using primarily binary operations. This results in 58x faster convolutional operations and 32x memory savings. XNOR-Nets offer the possibility of running state-of-the-art networks on CPUs (rather than GPUs) in real-time. Our binary networks are simple, accurate, efficient, and work on challenging visual tasks. We evaluate our approach on the ImageNet classification task. The classification accuracy with a Binary-Weight-Network version of AlexNet is only 2.9% less than the full-precision AlexNet (in top-1 measure). We compare our method with recent network binarization methods, BinaryConnect and BinaryNets, and outperform these methods by large margins on ImageNet, more than 16% in top-1 accuracy.}, note={arXiv:1603.05279}, number={arXiv:1603.05279}, publisher={arXiv}, author={Rastegari, Mohammad and Ordonez, Vicente and Redmon, Joseph and Farhadi, Ali}, year={2016}, month=aug }

@article{BiTAT_Park_2022, title={BiTAT: Neural Network Binarization with Task-dependent Aggregated Transformation}, url={http://arxiv.org/abs/2207.01394}, DOI={10.48550/arXiv.2207.01394}, abstractNote={Neural network quantization aims to transform high-precision weights and activations of a given neural network into low-precision weights/activations for reduced memory usage and computation, while preserving the performance of the original model. However, extreme quantization (1-bit weight/1-bit activations) of compactly-designed backbone architectures (e.g., MobileNets) often used for edge-device deployments results in severe performance degeneration. This paper proposes a novel Quantization-Aware Training (QAT) method that can effectively alleviate performance degeneration even with extreme quantization by focusing on the inter-weight dependencies, between the weights within each layer and across consecutive layers. To minimize the quantization impact of each weight on others, we perform an orthonormal transformation of the weights at each layer by training an input-dependent correlation matrix and importance vector, such that each weight is disentangled from the others. Then, we quantize the weights based on their importance to minimize the loss of the information from the original weights/activations. We further perform progressive layer-wise quantization from the bottom layer to the top, so that quantization at each layer reflects the quantized distributions of weights and activations at previous layers. We validate the effectiveness of our method on various benchmark datasets against strong neural quantization baselines, demonstrating that it alleviates the performance degeneration on ImageNet and successfully preserves the full-precision model performance on CIFAR-100 with compact backbone networks.}, note={arXiv:2207.01394}, number={arXiv:2207.01394}, publisher={arXiv}, author={Park, Geon and Yoon, Jaehong and Zhang, Haiyang and Zhang, Xing and Hwang, Sung Ju and Eldar, Yonina C.}, year={2022}, month=july }

@article{BiReal_Liu_2018, title={Bi-Real Net: Enhancing the Performance of 1-bit CNNs With Improved Representational Capability and Advanced Training Algorithm}, url={http://arxiv.org/abs/1808.00278}, DOI={10.48550/arXiv.1808.00278}, abstractNote={In this work, we study the 1-bit convolutional neural networks (CNNs), of which both the weights and activations are binary. While being efficient, the classification accuracy of the current 1-bit CNNs is much worse compared to their counterpart real-valued CNN models on the large-scale dataset, like ImageNet. To minimize the performance gap between the 1-bit and real-valued CNN models, we propose a novel model, dubbed Bi-Real net, which connects the real activations (after the 1-bit convolution and/or BatchNorm layer, before the sign function) to activations of the consecutive block, through an identity shortcut. Consequently, compared to the standard 1-bit CNN, the representational capability of the Bi-Real net is significantly enhanced and the additional cost on computation is negligible. Moreover, we develop a specific training algorithm including three technical novelties for 1- bit CNNs. Firstly, we derive a tight approximation to the derivative of the non-differentiable sign function with respect to activation. Secondly, we propose a magnitude-aware gradient with respect to the weight for updating the weight parameters. Thirdly, we pre-train the real-valued CNN model with a clip function, rather than the ReLU function, to better initialize the Bi-Real net. Experiments on ImageNet show that the Bi-Real net with the proposed training algorithm achieves 56.4% and 62.2% top-1 accuracy with 18 layers and 34 layers, respectively. Compared to the state-of-the-arts (e.g., XNOR Net), Bi-Real net achieves up to 10% higher top-1 accuracy with more memory saving and lower computational cost. Keywords: binary neural network, 1-bit CNNs, 1-layer-per-block}, note={arXiv:1808.00278}, number={arXiv:1808.00278}, publisher={arXiv}, author={Liu, Zechun and Wu, Baoyuan and Luo, Wenhan and Yang, Xin and Liu, Wei and Cheng, Kwang-Ting}, year={2018}, month=sept }

@article{BiPer_Vargas_2024a, title={BiPer: Binary Neural Networks using a Periodic Function}, url={http://arxiv.org/abs/2404.01278}, DOI={10.48550/arXiv.2404.01278}, abstractNote={Quantized neural networks employ reduced precision representations for both weights and activations. This quantization process significantly reduces the memory requirements and computational complexity of the network. Binary Neural Networks (BNNs) are the extreme quantization case, representing values with just one bit. Since the sign function is typically used to map real values to binary values, smooth approximations are introduced to mimic the gradients during error backpropagation. Thus, the mismatch between the forward and backward models corrupts the direction of the gradient, causing training inconsistency problems and performance degradation. In contrast to current BNN approaches, we propose to employ a binary periodic (BiPer) function during binarization. Specifically, we use a square wave for the forward pass to obtain the binary values and employ the trigonometric sine function with the same period of the square wave as a differentiable surrogate during the backward pass. We demonstrate that this approach can control the quantization error by using the frequency of the periodic function and improves network performance. Extensive experiments validate the effectiveness of BiPer in benchmark datasets and network architectures, with improvements of up to 1% and 0.69% with respect to state-of-the-art methods in the classification task over CIFAR-10 and ImageNet, respectively. Our code is publicly available at https://github.com/edmav4/BiPer.}, note={arXiv:2404.01278}, number={arXiv:2404.01278}, publisher={arXiv}, author={Vargas, Edwin and Correa, Claudia and Hinojosa, Carlos and Arguello, Henry}, year={2024}, month=apr }

@article{IR_Net_2020, title={Forward and Backward Information Retention for Accurate Binary Neural Networks}, url={http://arxiv.org/abs/1909.10788}, DOI={10.48550/arXiv.1909.10788}, abstractNote={Weight and activation binarization is an effective approach to deep neural network compression and can accelerate the inference by leveraging bitwise operations. Although many binarization methods have improved the accuracy of the model by minimizing the quantization error in forward propagation, there remains a noticeable performance gap between the binarized model and the full-precision one. Our empirical study indicates that the quantization brings information loss in both forward and backward propagation, which is the bottleneck of training accurate binary neural networks. To address these issues, we propose an Information Retention Network (IR-Net) to retain the information that consists in the forward activations and backward gradients. IR-Net mainly relies on two technical contributions: (1) Libra Parameter Binarization (Libra-PB): simultaneously minimizing both quantization error and information loss of parameters by balanced and standardized weights in forward propagation; (2) Error Decay Estimator (EDE): minimizing the information loss of gradients by gradually approximating the sign function in backward propagation, jointly considering the updating ability and accurate gradients. We are the first to investigate both forward and backward processes of binary networks from the unified information perspective, which provides new insight into the mechanism of network binarization. Comprehensive experiments with various network structures on CIFAR-10 and ImageNet datasets manifest that the proposed IR-Net can consistently outperform state-of-the-art quantization methods.}, note={arXiv:1909.10788}, number={arXiv:1909.10788}, publisher={arXiv}, author={Qin, Haotong and Gong, Ruihao and Liu, Xianglong and Shen, Mingzhu and Wei, Ziran and Yu, Fengwei and Song, Jingkuan}, year={2020}, month=mar }

@misc{XNOR_Plusplus_Bulat_2019, title={XNOR-Net++: Improved Binary Neural Networks}, url={https://arxiv.org/abs/1909.13863v1}, abstractNote={This paper proposes an improved training algorithm for binary neural networks in which both weights and activations are binary numbers. A key but fairly overlooked feature of the current state-of-the-art method of XNOR-Net is the use of analytically calculated real-valued scaling factors for re-weighting the output of binary convolutions. We argue that analytic calculation of these factors is sub-optimal. Instead, in this work, we make the following contributions: (a) we propose to fuse the activation and weight scaling factors into a single one that is learned discriminatively via backpropagation. (b) More importantly, we explore several ways of constructing the shape of the scale factors while keeping the computational budget fixed. (c) We empirically measure the accuracy of our approximations and show that they are significantly more accurate than the analytically calculated one. (d) We show that our approach significantly outperforms XNOR-Net within the same computational budget when tested on the challenging task of ImageNet classification, offering up to 6% accuracy gain.}, journal={arXiv.org}, author={Bulat, Adrian and Tzimiropoulos, Georgios}, year={2019}, month=sept, language={en} }

@article{Stochastic_Quantization_Dong_2017, title={Learning Accurate Low-Bit Deep Neural Networks with Stochastic Quantization}, url={http://arxiv.org/abs/1708.01001}, DOI={10.48550/arXiv.1708.01001}, abstractNote={Low-bit deep neural networks (DNNs) become critical for embedded applications due to their low storage requirement and computing efficiency. However, they suffer much from the non-negligible accuracy drop. This paper proposes the stochastic quantization (SQ) algorithm for learning accurate low-bit DNNs. The motivation is due to the following observation. Existing training algorithms approximate the real-valued elements/filters with low-bit representation all together in each iteration. The quantization errors may be small for some elements/filters, while are remarkable for others, which lead to inappropriate gradient direction during training, and thus bring notable accuracy drop. Instead, SQ quantizes a portion of elements/filters to low-bit with a stochastic probability inversely proportional to the quantization error, while keeping the other portion unchanged with full-precision. The quantized and full-precision portions are updated with corresponding gradients separately in each iteration. The SQ ratio is gradually increased until the whole network is quantized. This procedure can greatly compensate the quantization error and thus yield better accuracy for low-bit DNNs. Experiments show that SQ can consistently and significantly improve the accuracy for different low-bit DNNs on various datasets and various network structures.}, note={arXiv:1708.01001}, number={arXiv:1708.01001}, publisher={arXiv}, author={Dong, Yinpeng and Ni, Renkun and Li, Jianguo and Chen, Yurong and Zhu, Jun and Su, Hang}, year={2017}, month=aug }

@inproceedings{OvSW_Xiang, address={Cham}, title={OvSW: Overcoming Silent Weights for Accurate Binary Neural Networks}, ISBN={9783031734144}, DOI={10.1007/978-3-031-73414-4_1}, abstractNote={Binary Neural Networks (BNNs) have been proven to be highly effective for deploying deep neural networks on mobile and embedded platforms. Most existing works focus on minimizing quantization errors, improving representation ability, or designing gradient approximations to alleviate gradient mismatch in BNNs, while leaving the weight sign flipping, a critical factor for achieving powerful BNNs, untouched. In this paper, we investigate the efficiency of weight sign updates in BNNs. We observe that, for vanilla BNNs, over 50% of the weights remain their signs unchanged during training, and these weights are not only distributed at the tails of the weight distribution but also universally present in the vicinity of zero. We refer to these weights as “silent weights”, which slow down convergence and lead to a significant accuracy degradation. Theoretically, we reveal this is due to the independence of the BNNs gradient from the latent weight distribution. To address the issue, we propose Overcome Silent Weights (OvSW). OvSW first employs Adaptive Gradient Scaling (AGS) to establish a relationship between the gradient and the latent weight distribution, thereby improving the overall efficiency of weight sign updates. Additionally, we design Silence Awareness Decaying (SAD) to automatically identify “silent weights” by tracking weight flipping state, and apply an additional penalty to “silent weights” to facilitate their flipping. By efficiently updating weight signs, our method achieves faster convergence and state-of-the-art performance on CIFAR10 and ImageNet1K dataset with various architectures. For example, OvSW obtains 61.6% and 65.5% top-1 accuracy on the ImageNet1K using binarized ResNet18 and ResNet34 architecture respectively. Codes are available at https://github.com/JingyangXiang/OvSW.}, booktitle={Computer Vision – ECCV 2024}, publisher={Springer Nature Switzerland}, author={Xiang, Jingyang and Chen, Zuohui and Li, Siqi and Wu, Qing and Liu, Yong}, editor={Leonardis, Aleš and Ricci, Elisa and Roth, Stefan and Russakovsky, Olga and Sattler, Torsten and Varol, Gül}, year={2025}, pages={1–-18}, language={en} }

@inproceedings{ReCU_Xu_2021, title={ReCU: Reviving the Dead Weights in Binary Neural Networks}, ISSN={2380-7504}, url={https://ieeexplore.ieee.org/document/9710916}, DOI={10.1109/ICCV48922.2021.00515}, abstractNote={Binary neural networks (BNNs) have received increasing attention due to their superior reductions of computation and memory. Most existing works focus on either lessening the quantization error by minimizing the gap between the full-precision weights and their binarization or designing a gradient approximation to mitigate the gradient mismatch, while leaving the “dead weights” untouched. This leads to slow convergence when training BNNs. In this paper, for the first time, we explore the influence of “dead weights” which refer to a group of weights that are barely updated during the training of BNNs, and then introduce rectified clamp unit (ReCU) to revive the “dead weights” for updating. We prove that reviving the “dead weights” by ReCU can result in a smaller quantization error. Besides, we also take into account the information entropy of the weights, and then mathematically analyze why the weight standardization can benefit BNNs. We demonstrate the inherent contradiction between minimizing the quantization error and maximizing the information entropy, and then propose an adaptive exponential scheduler to identify the range of the “dead weights”. By considering the “dead weights”, our method offers not only faster BNN training, but also state-of-the-art performance on CIFAR-10 and ImageNet, compared with recent methods. Code can be available at https://github.com/z-hXu/ReCU.}, booktitle={2021 IEEE/CVF International Conference on Computer Vision (ICCV)}, author={Xu, Zihan and Lin, Mingbao and Liu, Jianzhuang and Chen, Jie and Shao, Ling and Gao, Yue and Tian, Yonghong and Ji, Rongrong}, year={2021}, month=oct, pages={5178–-5188} }

@article{ADMM_NN_Ren_2018, title={ADMM-NN: An Algorithm-Hardware Co-Design Framework of DNNs Using Alternating Direction Method of Multipliers}, url={http://arxiv.org/abs/1812.11677}, DOI={10.48550/arXiv.1812.11677}, abstractNote={To facilitate efficient embedded and hardware implementations of deep neural networks (DNNs), two important categories of DNN model compression techniques: weight pruning and weight quantization are investigated. The former leverages the redundancy in the number of weights, whereas the latter leverages the redundancy in bit representation of weights. However, there lacks a systematic framework of joint weight pruning and quantization of DNNs, thereby limiting the available model compression ratio. Moreover, the computation reduction, energy efficiency improvement, and hardware performance overhead need to be accounted for besides simply model size reduction. To address these limitations, we present ADMM-NN, the first algorithm-hardware co-optimization framework of DNNs using Alternating Direction Method of Multipliers (ADMM), a powerful technique to deal with non-convex optimization problems with possibly combinatorial constraints. The first part of ADMM-NN is a systematic, joint framework of DNN weight pruning and quantization using ADMM. It can be understood as a smart regularization technique with regularization target dynamically updated in each ADMM iteration, thereby resulting in higher performance in model compression than prior work. The second part is hardware-aware DNN optimizations to facilitate hardware-level implementations. Without accuracy loss, we can achieve 85$times$ and 24$times$ pruning on LeNet-5 and AlexNet models, respectively, significantly higher than prior work. The improvement becomes more significant when focusing on computation reductions. Combining weight pruning and quantization, we achieve 1,910$times$ and 231$times$ reductions in overall model size on these two benchmarks, when focusing on data storage. Highly promising results are also observed on other representative DNNs such as VGGNet and ResNet-50.}, note={arXiv:1812.11677}, number={arXiv:1812.11677}, publisher={arXiv}, author={Ren, Ao and Zhang, Tianyun and Ye, Shaokai and Li, Jiayu and Xu, Wenyao and Qian, Xuehai and Lin, Xue and Wang, Yanzhi}, year={2018}, month=dec }

@article{LQ_Net_Zhang_2018, title={LQ-Nets: Learned Quantization for Highly Accurate and Compact Deep Neural Networks}, url={http://arxiv.org/abs/1807.10029}, DOI={10.48550/arXiv.1807.10029}, abstractNote={Although weight and activation quantization is an effective approach for Deep Neural Network (DNN) compression and has a lot of potentials to increase inference speed leveraging bit-operations, there is still a noticeable gap in terms of prediction accuracy between the quantized model and the full-precision model. To address this gap, we propose to jointly train a quantized, bit-operation-compatible DNN and its associated quantizers, as opposed to using fixed, handcrafted quantization schemes such as uniform or logarithmic quantization. Our method for learning the quantizers applies to both network weights and activations with arbitrary-bit precision, and our quantizers are easy to train. The comprehensive experiments on CIFAR-10 and ImageNet datasets show that our method works consistently well for various network structures such as AlexNet, VGG-Net, GoogLeNet, ResNet, and DenseNet, surpassing previous quantization methods in terms of accuracy by an appreciable margin. Code available at https://github.com/Microsoft/LQ-Nets}, note={arXiv:1807.10029}, number={arXiv:1807.10029}, publisher={arXiv}, author={Zhang, Dongqing and Yang, Jiaolong and Ye, Dongqiangzi and Hua, Gang}, year={2018}, month=july }

@article{PACT_Choi_2018, title={PACT: Parameterized Clipping Activation for Quantized Neural Networks}, url={http://arxiv.org/abs/1805.06085}, DOI={10.48550/arXiv.1805.06085}, abstractNote={Deep learning algorithms achieve high classification accuracy at the expense of significant computation cost. To address this cost, a number of quantization schemes have been proposed - but most of these techniques focused on quantizing weights, which are relatively smaller in size compared to activations. This paper proposes a novel quantization scheme for activations during training - that enables neural networks to work well with ultra low precision weights and activations without any significant accuracy degradation. This technique, PArameterized Clipping acTivation (PACT), uses an activation clipping parameter $α$ that is optimized during training to find the right quantization scale. PACT allows quantizing activations to arbitrary bit precisions, while achieving much better accuracy relative to published state-of-the-art quantization schemes. We show, for the first time, that both weights and activations can be quantized to 4-bits of precision while still achieving accuracy comparable to full precision networks across a range of popular models and datasets. We also show that exploiting these reduced-precision computational units in hardware can enable a super-linear improvement in inferencing performance due to a significant reduction in the area of accelerator compute engines coupled with the ability to retain the quantized model and activation data in on-chip memories.}, note={arXiv:1805.06085}, number={arXiv:1805.06085}, publisher={arXiv}, author={Choi, Jungwook and Wang, Zhuo and Venkataramani, Swagath and Chuang, Pierce I.-Jen and Srinivasan, Vijayalakshmi and Gopalakrishnan, Kailash}, year={2018}, month=july }

@article{LSQ_Esser_2020, title={Learned Step Size Quantization}, url={http://arxiv.org/abs/1902.08153}, DOI={10.48550/arXiv.1902.08153}, abstractNote={Deep networks run with low precision operations at inference time offer power and space advantages over high precision alternatives, but need to overcome the challenge of maintaining high accuracy as precision decreases. Here, we present a method for training such networks, Learned Step Size Quantization, that achieves the highest accuracy to date on the ImageNet dataset when using models, from a variety of architectures, with weights and activations quantized to 2-, 3- or 4-bits of precision, and that can train 3-bit models that reach full precision baseline accuracy. Our approach builds upon existing methods for learning weights in quantized networks by improving how the quantizer itself is configured. Specifically, we introduce a novel means to estimate and scale the task loss gradient at each weight and activation layer’s quantizer step size, such that it can be learned in conjunction with other network parameters. This approach works using different levels of precision as needed for a given system and requires only a simple modification of existing training code.}, note={arXiv:1902.08153}, number={arXiv:1902.08153}, publisher={arXiv}, author={Esser, Steven K. and McKinstry, Jeffrey L. and Bablani, Deepika and Appuswamy, Rathinakumar and Modha, Dharmendra S.}, year={2020}, month=may }

@article{DoReFa_Zhou_2018, title={DoReFa-Net: Training Low Bitwidth Convolutional Neural Networks with Low Bitwidth Gradients}, url={http://arxiv.org/abs/1606.06160}, DOI={10.48550/arXiv.1606.06160}, abstractNote={We propose DoReFa-Net, a method to train convolutional neural networks that have low bitwidth weights and activations using low bitwidth parameter gradients. In particular, during backward pass, parameter gradients are stochastically quantized to low bitwidth numbers before being propagated to convolutional layers. As convolutions during forward/backward passes can now operate on low bitwidth weights and activations/gradients respectively, DoReFa-Net can use bit convolution kernels to accelerate both training and inference. Moreover, as bit convolutions can be efficiently implemented on CPU, FPGA, ASIC and GPU, DoReFa-Net opens the way to accelerate training of low bitwidth neural network on these hardware. Our experiments on SVHN and ImageNet datasets prove that DoReFa-Net can achieve comparable prediction accuracy as 32-bit counterparts. For example, a DoReFa-Net derived from AlexNet that has 1-bit weights, 2-bit activations, can be trained from scratch using 6-bit gradients to get 46.1% top-1 accuracy on ImageNet validation set. The DoReFa-Net AlexNet model is released publicly.}, note={arXiv:1606.06160}, number={arXiv:1606.06160}, publisher={arXiv}, author={Zhou, Shuchang and Wu, Yuxin and Ni, Zekun and Zhou, Xinyu and Wen, He and Zou, Yuheng}, year={2018}, month=feb }

@article{Oscillations_2022, title={Overcoming Oscillations in Quantization-Aware Training}, url={http://arxiv.org/abs/2203.11086}, DOI={10.48550/arXiv.2203.11086}, abstractNote={When training neural networks with simulated quantization, we observe that quantized weights can, rather unexpectedly, oscillate between two grid-points. The importance of this effect and its impact on quantization-aware training (QAT) are not well-understood or investigated in literature. In this paper, we delve deeper into the phenomenon of weight oscillations and show that it can lead to a significant accuracy degradation due to wrongly estimated batch-normalization statistics during inference and increased noise during training. These effects are particularly pronounced in low-bit ($leq$ 4-bits) quantization of efficient networks with depth-wise separable layers, such as MobileNets and EfficientNets. In our analysis we investigate several previously proposed QAT algorithms and show that most of these are unable to overcome oscillations. Finally, we propose two novel QAT algorithms to overcome oscillations during training: oscillation dampening and iterative weight freezing. We demonstrate that our algorithms achieve state-of-the-art accuracy for low-bit (3 & 4 bits) weight and activation quantization of efficient architectures, such as MobileNetV2, MobileNetV3, and EfficentNet-lite on ImageNet. Our source code is available at {https://github.com/qualcomm-ai-research/oscillations-qat}.}, note={arXiv:2203.11086}, number={arXiv:2203.11086}, publisher={arXiv}, author={Nagel, Markus and Fournarakis, Marios and Bondarenko, Yelysei and Blankevoort, Tijmen}, year={2022}, month=june }

@article{QNN_Hubara_2016, title={Quantized Neural Networks: Training Neural Networks with Low Precision Weights and Activations}, url={http://arxiv.org/abs/1609.07061}, DOI={10.48550/arXiv.1609.07061}, abstractNote={We introduce a method to train Quantized Neural Networks (QNNs) --- neural networks with extremely low precision (e.g., 1-bit) weights and activations, at run-time. At train-time the quantized weights and activations are used for computing the parameter gradients. During the forward pass, QNNs drastically reduce memory size and accesses, and replace most arithmetic operations with bit-wise operations. As a result, power consumption is expected to be drastically reduced. We trained QNNs over the MNIST, CIFAR-10, SVHN and ImageNet datasets. The resulting QNNs achieve prediction accuracy comparable to their 32-bit counterparts. For example, our quantized version of AlexNet with 1-bit weights and 2-bit activations achieves $51%$ top-1 accuracy. Moreover, we quantize the parameter gradients to 6-bits as well which enables gradients computation using only bit-wise operation. Quantized recurrent neural networks were tested over the Penn Treebank dataset, and achieved comparable accuracy as their 32-bit counterparts using only 4-bits. Last but not least, we programmed a binary matrix multiplication GPU kernel with which it is possible to run our MNIST QNN 7 times faster than with an unoptimized GPU kernel, without suffering any loss in classification accuracy. The QNN code is available online.}, note={arXiv:1609.07061}, number={arXiv:1609.07061}, publisher={arXiv}, author={Hubara, Itay and Courbariaux, Matthieu and Soudry, Daniel and El-Yaniv, Ran and Bengio, Yoshua}, year={2016}, month=sept }

@article{Yin_Zhang_Lyu_Osher_Qi_Xin_2018, title={BinaryRelax: A Relaxation Approach For Training Deep Neural Networks With Quantized Weights}, url={http://arxiv.org/abs/1801.06313}, DOI={10.48550/arXiv.1801.06313}, abstractNote={We propose BinaryRelax, a simple two-phase algorithm, for training deep neural networks with quantized weights. The set constraint that characterizes the quantization of weights is not imposed until the late stage of training, and a sequence of emph{pseudo} quantized weights is maintained. Specifically, we relax the hard constraint into a continuous regularizer via Moreau envelope, which turns out to be the squared Euclidean distance to the set of quantized weights. The pseudo quantized weights are obtained by linearly interpolating between the float weights and their quantizations. A continuation strategy is adopted to push the weights towards the quantized state by gradually increasing the regularization parameter. In the second phase, exact quantization scheme with a small learning rate is invoked to guarantee fully quantized weights. We test BinaryRelax on the benchmark CIFAR and ImageNet color image datasets to demonstrate the superiority of the relaxed quantization approach and the improved accuracy over the state-of-the-art training methods. Finally, we prove the convergence of BinaryRelax under an approximate orthogonality condition.}, note={arXiv:1801.06313}, number={arXiv:1801.06313}, publisher={arXiv}, author={Yin, Penghang and Zhang, Shuai and Lyu, Jiancheng and Osher, Stanley and Qi, Yingyong and Xin, Jack}, year={2018}, month=sept }

@article{Yuan_Agaian_2023, title={A comprehensive review of Binary Neural Network}, volume={56}, ISSN={1573-7462}, url={https://doi.org/10.1007/s10462-023-10464-w}, DOI={10.1007/s10462-023-10464-w}, abstractNote={Deep learning (DL) has recently changed the development of intelligent systems and is widely adopted in many real-life applications. Despite their various benefits and potentials, there is a high demand for DL processing in different computationally limited and energy-constrained devices. It is natural to study game-changing technologies such as Binary Neural Networks (BNN) to increase DL capabilities. Recently remarkable progress has been made in BNN since they can be implemented and embedded on tiny restricted devices and save a significant amount of storage, computation cost, and energy consumption. However, nearly all BNN acts trade with extra memory, computation cost, and higher performance. This article provides a complete overview of recent developments in BNN. This article focuses exclusively on 1-bit activations and weights 1-bit convolution networks, contrary to previous surveys in which low-bit works are mixed in. It conducted a complete investigation of BNN’s development—from their predecessors to the latest BNN algorithms/techniques, presenting a broad design pipeline and discussing each module’s variants. Along the way, it examines BNN (a) purpose: their early successes and challenges; (b) BNN optimization: selected representative works that contain essential optimization techniques; (c) deployment: open-source frameworks for BNN modeling and development; (d) terminal: efficient computing architectures and devices for BNN and (e) applications: diverse applications with BNN. Moreover, this paper discusses potential directions and future research opportunities in each section.}, number={11}, journal={Artificial Intelligence Review}, author={Yuan, Chunyu and Agaian, Sos S.}, year={2023}, month=nov, pages={12949–-13013}, language={en} }

@article{INQ_Zhou_2017a, title={Incremental Network Quantization: Towards Lossless CNNs with Low-Precision Weights}, url={http://arxiv.org/abs/1702.03044}, DOI={10.48550/arXiv.1702.03044}, abstractNote={This paper presents incremental network quantization (INQ), a novel method, targeting to efficiently convert any pre-trained full-precision convolutional neural network (CNN) model into a low-precision version whose weights are constrained to be either powers of two or zero. Unlike existing methods which are struggled in noticeable accuracy loss, our INQ has the potential to resolve this issue, as benefiting from two innovations. On one hand, we introduce three interdependent operations, namely weight partition, group-wise quantization and re-training. A well-proven measure is employed to divide the weights in each layer of a pre-trained CNN model into two disjoint groups. The weights in the first group are responsible to form a low-precision base, thus they are quantized by a variable-length encoding method. The weights in the other group are responsible to compensate for the accuracy loss from the quantization, thus they are the ones to be re-trained. On the other hand, these three operations are repeated on the latest re-trained group in an iterative manner until all the weights are converted into low-precision ones, acting as an incremental network quantization and accuracy enhancement procedure. Extensive experiments on the ImageNet classification task using almost all known deep CNN architectures including AlexNet, VGG-16, GoogleNet and ResNets well testify the efficacy of the proposed method. Specifically, at 5-bit quantization, our models have improved accuracy than the 32-bit floating-point references. Taking ResNet-18 as an example, we further show that our quantized models with 4-bit, 3-bit and 2-bit ternary weights have improved or very similar accuracy against its 32-bit floating-point baseline. Besides, impressive results with the combination of network pruning and INQ are also reported. The code is available at https://github.com/Zhouaojun/Incremental-Network-Quantization.}, note={arXiv:1702.03044}, number={arXiv:1702.03044}, publisher={arXiv}, author={Zhou, Aojun and Yao, Anbang and Guo, Yiwen and Xu, Lin and Chen, Yurong}, year={2017}, month=aug }

@inproceedings{lin2017towards,
  title={Towards Accurate Binary Convolutional Neural Network},
  author={Lin, Xiaofan and Zhao, Cong and Pan, Wei},
  booktitle={Advances in Neural Information Processing Systems},
  pages={345--353},
  year={2017}
}

@inproceedings{bethge2021meliusnet,
  title={MeliusNet: An Improved Network Architecture for Binary Neural Networks},
  author={Bethge, Joseph and Bartz, Christian and Yang, Haojin and Chen, Ying and Meinel, Christoph},
  booktitle={Proceedings of the IEEE/CVF Winter Conference on Applications of Computer Vision},
  pages={1600--1609},
  year={2021}
}

@inproceedings{leroux2020training,
  title={Training Binary Neural Networks with Knowledge Transfer},
  author={Leroux, Sam and Vankeirsbilck, Bert and Verbelen, Tim and Simoens, Pieter},
  booktitle={Neurocomputing},
  volume={396},
  pages={534--541},
  year={2020}
}

@inproceedings{liu2021adam,
  title={How Do Adam and Training Strategies Help BNNs Optimization?},
  author={Liu, Zechun and Shen, Zhiqiang and Li, Shichao and Helwegen, Koen and Huang, Dong and Cheng, Kwang-Ting},
  booktitle={Proceedings of the 38th International Conference on Machine Learning},
  pages={6936--6946},
  year={2021}
}

@inproceedings{liu2020reactnet,
  title={ReActNet: Towards Precise Binary Neural Network with Generalized Activation Functions},
  author={Liu, Zechun and Shen, Zhiqiang and Savvides, Marios and Cheng, Kwang-Ting},
  booktitle={European Conference on Computer Vision},
  pages={143--159},
  year={2020}
}

@inproceedings{kim2021improving,
  title={Improving Accuracy of Binary Neural Networks using Unbalanced Activation Distribution},
  author={Kim, Hyungjun and Kim, Yeshwanth and Choi, Sungjoo and Yoo, Hoi-Jun},
  booktitle={Proceedings of the IEEE/CVF Conference on Computer Vision and Pattern Recognition},
  pages={7862--7871},
  year={2021}
}

@misc{tu2022adabinimprovingbinaryneural,
      title={AdaBin: Improving Binary Neural Networks with Adaptive Binary Sets}, 
      author={Zhijun Tu and Xinghao Chen and Pengju Ren and Yunhe Wang},
      year={2022},
      eprint={2208.08084},
      archivePrefix={arXiv},
      primaryClass={cs.CV},
      url={https://arxiv.org/abs/2208.08084}, 
}

@article{QKD_Kim_2019, title={QKD: Quantization-aware Knowledge Distillation}, journal=ArXiv, url={http://arxiv.org/abs/1911.12491}, DOI={10.48550/arXiv.1911.12491}, abstractNote={Quantization and Knowledge distillation (KD) methods are widely used to reduce memory and power consumption of deep neural networks (DNNs), especially for resource-constrained edge devices. Although their combination is quite promising to meet these requirements, it may not work as desired. It is mainly because the regularization effect of KD further diminishes the already reduced representation power of a quantized model. To address this short-coming, we propose Quantization-aware Knowledge Distillation (QKD) wherein quantization and KD are care-fully coordinated in three phases. First, Self-studying (SS) phase fine-tunes a quantized low-precision student network without KD to obtain a good initialization. Second, Co-studying (CS) phase tries to train a teacher to make it more quantizaion-friendly and powerful than a fixed teacher. Finally, Tutoring (TU) phase transfers knowledge from the trained teacher to the student. We extensively evaluate our method on ImageNet and CIFAR-10/100 datasets and show an ablation study on networks with both standard and depthwise-separable convolutions. The proposed QKD outperformed existing state-of-the-art methods (e.g., 1.3% improvement on ResNet-18 with W4A4, 2.6% on MobileNetV2 with W4A4). Additionally, QKD could recover the full-precision accuracy at as low as W3A3 quantization on ResNet and W6A6 quantization on MobilenetV2.}, note={arXiv:1911.12491}, number={arXiv:1911.12491}, publisher={arXiv}, author={Kim, Jangho and Bhalgat, Yash and Lee, Jinwon and Patel, Chirag and Kwak, Nojun}, year={2019}, month=nov }

@article{gao2022memristive,
title = {Memristive KDG-BNN: Memristive binary neural networks trained via knowledge distillation and generative adversarial networks},
journal = {Knowledge-Based Systems},
volume = {249},
pages = {108962},
year = {2022},
issn = {0950-7051},
doi = {https://doi.org/10.1016/j.knosys.2022.108962},
url = {https://www.sciencedirect.com/science/article/pii/S095070512200466X},
author = {Tongtong Gao and Yue Zhou and Shukai Duan and Xiaofang Hu}
}

@misc{liu2019circulantbinaryconvolutionalnetworks,
      title={Circulant Binary Convolutional Networks: Enhancing the Performance of 1-bit DCNNs with Circulant Back Propagation}, 
      author={Chunlei Liu and Wenrui Ding and Xin Xia and Baochang Zhang and Jiaxin Gu and Jianzhuang Liu and Rongrong Ji and David Doermann},
      year={2019},
      eprint={1910.10853},
      archivePrefix={arXiv},
      primaryClass={cs.CV},
      url={https://arxiv.org/abs/1910.10853}, 
}

@article{Bethge2019BinaryDenseNetDA,
  title={BinaryDenseNet: Developing an Architecture for Binary Neural Networks},
  author={Joseph Bethge and Haojin Yang and Marvin Bornstein and Christoph Meinel},
  journal={2019 IEEE/CVF International Conference on Computer Vision Workshop (ICCVW)},
  year={2019},
  pages={1951--1960},
  url={https://api.semanticscholar.org/CorpusID:207901458}
}

@misc{zhuang2022structuredbinaryneuralnetworks,
      title={Structured Binary Neural Networks for Image Recognition}, 
      author={Bohan Zhuang and Chunhua Shen and Mingkui Tan and Peng Chen and Lingqiao Liu and Ian Reid},
      year={2022},
      eprint={1909.09934},
      archivePrefix={arXiv},
      primaryClass={cs.CV},
      url={https://arxiv.org/abs/1909.09934}, 
}

@misc{helwegen2019latentweightsexistrethinking,
      title={Latent Weights Do Not Exist: Rethinking Binarized Neural Network Optimization}, 
      author={Koen Helwegen and James Widdicombe and Lukas Geiger and Zechun Liu and Kwang-Ting Cheng and Roeland Nusselder},
      year={2019},
      eprint={1906.02107},
      archivePrefix={arXiv},
      primaryClass={cs.LG},
      url={https://arxiv.org/abs/1906.02107}, 
}

@misc{kim2020binaryduoreducinggradientmismatch,
      title={BinaryDuo: Reducing Gradient Mismatch in Binary Activation Network by Coupling Binary Activations}, 
      author={Hyungjun Kim and Kyungsu Kim and Jinseok Kim and Jae-Joon Kim},
      year={2020},
      eprint={2002.06517},
      archivePrefix={arXiv},
      primaryClass={cs.LG},
      url={https://arxiv.org/abs/2002.06517}, 
}

@misc{lin2019defensivequantizationefficiencymeets,
      title={Defensive Quantization: When Efficiency Meets Robustness}, 
      author={Ji Lin and Chuang Gan and Song Han},
      year={2019},
      eprint={1904.08444},
      archivePrefix={arXiv},
      primaryClass={cs.LG},
      url={https://arxiv.org/abs/1904.08444}, 
}

@InProceedings{hou2017loss,
	title={Loss-aware Binarization of Deep Networks},
	author={Hou, Lu and Yao, Quanming and Kwok, James T.},
	booktitle={International Conference on Learning Representations},
	year={2017}
}

@article{sgdat2023,
    author = {Shan, Gu and Guoyin, Zhang and Chengwei, Jia and Yanxia, Wu},
    title = {SGDAT: An optimization method for binary neural networks},
    year = {2023},
    issue_date = {Oct 2023},
    publisher = {Elsevier Science Publishers B. V.},
    address = {NLD},
    volume = {555},
    number = {C},
    issn = {0925-2312},
    url = {https://doi.org/10.1016/j.neucom.2023.126431},
    doi = {10.1016/j.neucom.2023.126431},
    journal = {Neurocomput.},
    month = oct,
    numpages = {8}
}

@misc{ding2019regularizingactivationdistributiontraining,
      title={Regularizing Activation Distribution for Training Binarized Deep Networks}, 
      author={Ruizhou Ding and Ting-Wu Chin and Zeye Liu and Diana Marculescu},
      year={2019},
      eprint={1904.02823},
      archivePrefix={arXiv},
      primaryClass={cs.CV},
      url={https://arxiv.org/abs/1904.02823}, 
}

@misc{gu2018projectionconvolutionalneuralnetworks,
      title={Projection Convolutional Neural Networks for 1-bit CNNs via Discrete Back Propagation}, 
      author={Jiaxin Gu and Ce Li and Baochang Zhang and Jungong Han and Xianbin Cao and Jianzhuang Liu and David Doermann},
      year={2018},
      eprint={1811.12755},
      archivePrefix={arXiv},
      primaryClass={cs.CV},
      url={https://arxiv.org/abs/1811.12755}, 
}

@misc{zhang2026sparsebitnet158bitllmsnaturally,
      title={Sparse-BitNet: 1.58-bit LLMs are Naturally Friendly to Semi-Structured Sparsity}, 
      author={Di Zhang and Xun Wu and Shaohan Huang and Yudong Wang and Hanyong Shao and Yingbo Hao and Zewen Chi and Li Dong and Ting Song and Yan Xia and Zhifang Sui and Furu Wei},
      year={2026},
      eprint={2603.05168},
      archivePrefix={arXiv},
      primaryClass={cs.CL},
      url={https://arxiv.org/abs/2603.05168}, 
}

@article{Wang_Ma_Dong_Huang_Wang_Ma_Yang_Wang_Wu_Wei_2023, title={BitNet: Scaling 1-bit Transformers for Large Language Models}, url={http://arxiv.org/abs/2310.11453}, DOI={10.48550/arXiv.2310.11453}, abstractNote={The increasing size of large language models has posed challenges for deployment and raised concerns about environmental impact due to high energy consumption. In this work, we introduce BitNet, a scalable and stable 1-bit Transformer architecture designed for large language models. Specifically, we introduce BitLinear as a drop-in replacement of the nn.Linear layer in order to train 1-bit weights from scratch. Experimental results on language modeling show that BitNet achieves competitive performance while substantially reducing memory footprint and energy consumption, compared to state-of-the-art 8-bit quantization methods and FP16 Transformer baselines. Furthermore, BitNet exhibits a scaling law akin to full-precision Transformers, suggesting its potential for effective scaling to even larger language models while maintaining efficiency and performance benefits.}, note={arXiv:2310.11453}, number={arXiv:2310.11453}, publisher={arXiv}, author={Wang, Hongyu and Ma, Shuming and Dong, Li and Huang, Shaohan and Wang, Huaijie and Ma, Lingxiao and Yang, Fan and Wang, Ruiping and Wu, Yi and Wei, Furu}, year={2023}, month=oct }

@article{Dosovitskiy_Beyer_Kolesnikov_Weissenborn_Zhai_Unterthiner_Dehghani_Minderer_Heigold_Gelly_et, title={An Image is Worth 16x16 Words: Transformers for Image Recognition at Scale}, url={http://arxiv.org/abs/2010.11929}, DOI={10.48550/arXiv.2010.11929}, abstractNote={While the Transformer architecture has become the de-facto standard for natural language processing tasks, its applications to computer vision remain limited. In vision, attention is either applied in conjunction with convolutional networks, or used to replace certain components of convolutional networks while keeping their overall structure in place. We show that this reliance on CNNs is not necessary and a pure transformer applied directly to sequences of image patches can perform very well on image classification tasks. When pre-trained on large amounts of data and transferred to multiple mid-sized or small image recognition benchmarks (ImageNet, CIFAR-100, VTAB, etc.), Vision Transformer (ViT) attains excellent results compared to state-of-the-art convolutional networks while requiring substantially fewer computational resources to train.}, note={arXiv:2010.11929}, number={arXiv:2010.11929}, publisher={arXiv}, author={Dosovitskiy, Alexey and Beyer, Lucas and Kolesnikov, Alexander and Weissenborn, Dirk and Zhai, Xiaohua and Unterthiner, Thomas and Dehghani, Mostafa and Minderer, Matthias and Heigold, Georg and Gelly, Sylvain and Uszkoreit, Jakob and Houlsby, Neil}, year={2021}, month=june }

@article{Bai_Zhang_Hou_Shang_Jin_Jiang_Liu_Lyu_King_2021, title={BinaryBERT: Pushing the Limit of BERT Quantization}, url={http://arxiv.org/abs/2012.15701}, DOI={10.48550/arXiv.2012.15701}, abstractNote={The rapid development of large pre-trained language models has greatly increased the demand for model compression techniques, among which quantization is a popular solution. In this paper, we propose BinaryBERT, which pushes BERT quantization to the limit by weight binarization. We find that a binary BERT is hard to be trained directly than a ternary counterpart due to its complex and irregular loss landscape. Therefore, we propose ternary weight splitting, which initializes BinaryBERT by equivalently splitting from a half-sized ternary network. The binary model thus inherits the good performance of the ternary one, and can be further enhanced by fine-tuning the new architecture after splitting. Empirical results show that our BinaryBERT has only a slight performance drop compared with the full-precision model while being 24x smaller, achieving the state-of-the-art compression results on the GLUE and SQuAD benchmarks.}, note={arXiv:2012.15701}, number={arXiv:2012.15701}, publisher={arXiv}, author={Bai, Haoli and Zhang, Wei and Hou, Lu and Shang, Lifeng and Jin, Jing and Jiang, Xin and Liu, Qun and Lyu, Michael and King, Irwin}, year={2021}, month=july }

@article{Devlin_Chang_Lee_Toutanova_2019, title={BERT: Pre-training of Deep Bidirectional Transformers for Language Understanding}, url={http://arxiv.org/abs/1810.04805}, DOI={10.48550/arXiv.1810.04805}, abstractNote={We introduce a new language representation model called BERT, which stands for Bidirectional Encoder Representations from Transformers. Unlike recent language representation models, BERT is designed to pre-train deep bidirectional representations from unlabeled text by jointly conditioning on both left and right context in all layers. As a result, the pre-trained BERT model can be fine-tuned with just one additional output layer to create state-of-the-art models for a wide range of tasks, such as question answering and language inference, without substantial task-specific architecture modifications. BERT is conceptually simple and empirically powerful. It obtains new state-of-the-art results on eleven natural language processing tasks, including pushing the GLUE score to 80.5% (7.7% point absolute improvement), MultiNLI accuracy to 86.7% (4.6% absolute improvement), SQuAD v1.1 question answering Test F1 to 93.2 (1.5 point absolute improvement) and SQuAD v2.0 Test F1 to 83.1 (5.1 point absolute improvement).}, note={arXiv:1810.04805}, number={arXiv:1810.04805}, publisher={arXiv}, author={Devlin, Jacob and Chang, Ming-Wei and Lee, Kenton and Toutanova, Kristina}, year={2019}, month=may }

@article{Zhang_Li_Liu_2023, title={On Uniform Scalar Quantization for Learned Image Compression}, url={http://arxiv.org/abs/2309.17051}, DOI={10.48550/arXiv.2309.17051}, abstractNote={Learned image compression possesses a unique challenge when incorporating non-differentiable quantization into the gradient-based training of the networks. Several quantization surrogates have been proposed to fulfill the training, but they were not systematically justified from a theoretical perspective. We fill this gap by contrasting uniform scalar quantization, the most widely used category with rounding being its simplest case, and its training surrogates. In principle, we find two factors crucial: one is the discrepancy between the surrogate and rounding, leading to train-test mismatch; the other is gradient estimation risk due to the surrogate, which consists of bias and variance of the gradient estimation. Our analyses and simulations imply that there is a tradeoff between the train-test mismatch and the gradient estimation risk, and the tradeoff varies across different network structures. Motivated by these analyses, we present a method based on stochastic uniform annealing, which has an adjustable temperature coefficient to control the tradeoff. Moreover, our analyses enlighten us as to two subtle tricks: one is to set an appropriate lower bound for the variance parameter of the estimated quantized latent distribution, which effectively reduces the train-test mismatch; the other is to use zero-center quantization with partial stop-gradient, which reduces the gradient estimation variance and thus stabilize the training. Our method with the tricks is verified to outperform the existing practices of quantization surrogates on a variety of representative image compression networks.}, note={arXiv:2309.17051}, number={arXiv:2309.17051}, publisher={arXiv}, author={Zhang, Haotian and Li, Li and Liu, Dong}, year={2023}, month=sept }

@article{Liu_Oguz_Pappu_Xiao_Yih_Li_Krishnamoorthi_Mehdad_2022, title={BiT: Robustly Binarized Multi-distilled Transformer}, url={http://arxiv.org/abs/2205.13016}, DOI={10.48550/arXiv.2205.13016}, abstractNote={Modern pre-trained transformers have rapidly advanced the state-of-the-art in machine learning, but have also grown in parameters and computational complexity, making them increasingly difficult to deploy in resource-constrained environments. Binarization of the weights and activations of the network can significantly alleviate these issues, however, is technically challenging from an optimization perspective. In this work, we identify a series of improvements that enables binary transformers at a much higher accuracy than what was possible previously. These include a two-set binarization scheme, a novel elastic binary activation function with learned parameters, and a method to quantize a network to its limit by successively distilling higher precision models into lower precision students. These approaches allow for the first time, fully binarized transformer models that are at a practical level of accuracy, approaching a full-precision BERT baseline on the GLUE language understanding benchmark within as little as 5.9%. Code and models are available at: https://github.com/facebookresearch/bit.}, note={arXiv:2205.13016}, number={arXiv:2205.13016}, publisher={arXiv}, author={Liu, Zechun and Oguz, Barlas and Pappu, Aasish and Xiao, Lin and Yih, Scott and Li, Meng and Krishnamoorthi, Raghuraman and Mehdad, Yashar}, year={2022}, month=oct }

@article{Qin_Ding_Zhang_Yan_Liu_Dang_Liu_Liu_2022, title={BiBERT: Accurate Fully Binarized BERT}, url={http://arxiv.org/abs/2203.06390}, DOI={10.48550/arXiv.2203.06390}, abstractNote={The large pre-trained BERT has achieved remarkable performance on Natural Language Processing (NLP) tasks but is also computation and memory expensive. As one of the powerful compression approaches, binarization extremely reduces the computation and memory consumption by utilizing 1-bit parameters and bitwise operations. Unfortunately, the full binarization of BERT (i.e., 1-bit weight, embedding, and activation) usually suffer a significant performance drop, and there is rare study addressing this problem. In this paper, with the theoretical justification and empirical analysis, we identify that the severe performance drop can be mainly attributed to the information degradation and optimization direction mismatch respectively in the forward and backward propagation, and propose BiBERT, an accurate fully binarized BERT, to eliminate the performance bottlenecks. Specifically, BiBERT introduces an efficient Bi-Attention structure for maximizing representation information statistically and a Direction-Matching Distillation (DMD) scheme to optimize the full binarized BERT accurately. Extensive experiments show that BiBERT outperforms both the straightforward baseline and existing state-of-the-art quantized BERTs with ultra-low bit activations by convincing margins on the NLP benchmark. As the first fully binarized BERT, our method yields impressive 56.3 times and 31.2 times saving on FLOPs and model size, demonstrating the vast advantages and potential of the fully binarized BERT model in real-world resource-constrained scenarios.}, note={arXiv:2203.06390}, number={arXiv:2203.06390}, publisher={arXiv}, author={Qin, Haotong and Ding, Yifu and Zhang, Mingyuan and Yan, Qinghua and Liu, Aishan and Dang, Qingqing and Liu, Ziwei and Liu, Xianglong}, year={2022}, month=mar }

@misc{GLUE_Wang_Singh_Michael_Hill_Levy_Bowman_2018, title={GLUE: A Multi-Task Benchmark and Analysis Platform for Natural Language Understanding}, url={https://arxiv.org/abs/1804.07461v3}, abstractNote={For natural language understanding (NLU) technology to be maximally useful, both practically and as a scientific object of study, it must be general: it must be able to process language in a way that is not exclusively tailored to any one specific task or dataset. In pursuit of this objective, we introduce the General Language Understanding Evaluation benchmark (GLUE), a tool for evaluating and analyzing the performance of models across a diverse range of existing NLU tasks. GLUE is model-agnostic, but it incentivizes sharing knowledge across tasks because certain tasks have very limited training data. We further provide a hand-crafted diagnostic test suite that enables detailed linguistic analysis of NLU models. We evaluate baselines based on current methods for multi-task and transfer learning and find that they do not immediately give substantial improvements over the aggregate performance of training a separate model per task, indicating room for improvement in developing general and robust NLU systems.}, journal={arXiv.org}, author={Wang, Alex and Singh, Amanpreet and Michael, Julian and Hill, Felix and Levy, Omer and Bowman, Samuel R.}, year={2018}, month=apr, language={en} }

@inproceedings{SST2_Socher2013RecursiveDM,
  title={Recursive Deep Models for Semantic Compositionality Over a Sentiment Treebank},
  author={Richard Socher and Alex Perelygin and Jean Wu and Jason Chuang and Christopher D. Manning and A. Ng and Christopher Potts},
  booktitle={Conference on Empirical Methods in Natural Language Processing},
  year={2013},
  url={https://api.semanticscholar.org/CorpusID:990233}
}

%%%%%%%%%%%%%%%%%%%%%%%%%%%%%%%%%%%%%%%%%%%%%%%%%%%%%%%%%%%%
\newpage
\appendix

\section{Training Recipes and Configurations}
\label{sec:appendix_a}

As described in the main paper in Section~\ref{sec:exp_setup}, we apply a minimal training condition with minimal modifications (no lr schedule, no weight decay) from a standard full precision schedule. Below are the precise training configurations for each experiments. We use default hyperparameters of $p(t)=(t/T)^3$ and refresh rate $r=100$.

\subsection{Main Results Configuration}
\label{sec:main_results_config}

Table~\ref{tab:training_config_cifar_main} provides the training configurations used for our main results (Table~\ref{tab:main_results}). We report settings for each dataset separately due to differences in resolution, augmentation, and training length. For our CIFAR-10 and CIFAR-100 results, we apply the same training recipe, apart from the dataset.

We apply a similar but marginally different training recipe to ImageNet as shown in Table~\ref{tab:training_config_imagenet_main}. We do this to align with standard full precision practices for ImageNet, again adhering to a minimal training recipe without schedulers or weight decay.

\subsection{Ablation Configuration}
\label{sec:ablation_configuration}
 Table~\ref{tab:experiment_overview} provides the overview of training scheme for the experimental ablations. All of our ablations use the same scheme aside from the controlled variable that is changed (for example, when sweeping epochs in Figure~\ref{fig:epoch_ablation}, this configuration was used only varying the epochs). We outline the modifications each ablation makes to the training scheme in Table~\ref{tab:training_config_ablations}.

\subsection{BiReal-Net Configuration}

For our BiReal-Net, train with a learning rate of $0.06$ over $300$ epochs. These particular values were selected to align with two general prinicples used for training BiReal-Net \citep{BiReal_Liu_2018}.

\begin{enumerate}[nosep]  
\item BiReal-Net uses a lower learning rate than standard ResNets, beginning at 0.01 throughout training.  
\item BiReal-Net is trained for an extended number of epochs when compared to a full precision ImageNet.
\end{enumerate}

For these reasons, we apply many epochs (300 for training from scratch, and an additional 300 for pretraining). We apply this scheme due to the discussed difficulties of applying StoMPP to learning rate schedules (\ref{sec:exp_setup}) and for simplicity. The pretrained networks for both StoMPP and STE use clip activation functions, but apply no freezing and no forward pass quantization, respectively. Aside from these changes ($lr=0.06$, 300 epochs), we use the training recipe described in the Ablation Configuration (\ref{sec:ablation_configuration}).

\begin{table}[H]
\centering
\caption{Training configuration for CIFAR-10 and CIFAR-100 main results (Table~\ref{tab:main_results}). Both datasets use identical settings except for the dataset itself.}
\label{tab:training_config_cifar_main}
\small
\begin{tabular}{ll}
\toprule
\textbf{Category} & \textbf{Setting} \\
\midrule
Dataset & CIFAR-10 / CIFAR-100 \\
Input resolution & $32 \times 32$ \\
Train batch size & 256 \\
Test batch size & 256 \\
Data augmentation & RandomCrop(32, pad=4), HorizontalFlip \\
Normalization & mean (0.5071, 0.4865, 0.4409), std (0.2673, 0.2564, 0.2762) \\
Model architectures & ResNet-18, ResNet-34, ResNet-50 \\
Model variants & STE-BNN, StoMPP-BNN, STE-BWN, StoMPP-BWN \\
Optimizer & SGD (Nesterov momentum) \\
Learning rate & 0.1 \\
Momentum & 0.9 \\
Weight decay & 0 \\
LR schedule & Constant \\
Loss & Cross-entropy \\
Label smoothing & 0.0 \\
Masking scheme (StoMPP) & Layerwise progressive masking \\
Mask schedule (StoMPP) & Cubic ($p: 0 \rightarrow 1$) \\
Mask refresh rate (StoMPP) & 100 \\
Activation function (unfrozen) & Clip \\
Frozen activation & Sign \\
Training epochs & 200 \\
Precision & FP32 \\
\bottomrule
\end{tabular}
\end{table}

\begin{table}[H]
\centering
\caption{Training configuration for ImageNet main results (Table~\ref{tab:main_results}). Training epochs vary by architecture as shown below.}
\label{tab:training_config_imagenet_main}
\small
\begin{tabular}{ll}
\toprule
\textbf{Category} & \textbf{Setting} \\
\midrule
Dataset & ImageNet (ILSVRC2012) \\
Input resolution & $224 \times 224$ \\
Train batch size & 256 \\
Test batch size & 256 \\
Data augmentation & RandomResizedCrop(224), HorizontalFlip \\
Normalization & mean (0.485, 0.456, 0.406), std (0.229, 0.224, 0.225) \\
Model architectures & ResNet-18, ResNet-34, ResNet-50 \\
\textbf{Training epochs} & ResNet-18: 97, ResNet-34: 131, ResNet-50: 98 \\
Model variants & STE-BNN, StoMPP-BNN, STE-BWN, StoMPP-BWN \\
Optimizer & SGD (Nesterov momentum) \\
Learning rate & 0.1 \\
Momentum & 0.9 \\
Weight decay & 0 \\
LR schedule & Constant \\
Loss & Cross-entropy \\
Label smoothing & 0.0 \\
Masking scheme (StoMPP) & Layerwise progressive masking \\
Mask schedule (StoMPP) & Cubic ($p: 0 \rightarrow 1$) \\
Mask refresh rate (StoMPP) & 100 \\
Activation function (unfrozen) & Clip \\
Frozen activation & Sign \\
Precision & FP8 \\
\bottomrule
\end{tabular}
\end{table}

% \begin{table}[h!]
% \centering
% \caption{Dataset-specific training details for main experiments.}
% \label{tab:dataset_training_overrides}
% \small
% \begin{tabular}{llll}
% \toprule
% \textbf{Dataset} & \textbf{Input} & \textbf{Epochs} & \textbf{Data augmentation} \\
% \midrule
% CIFAR-10 &
% $32 \times 32$ &
% 200 &
% RandomCrop(32, pad=4), HorizontalFlip \\

% CIFAR-100 &
% $32 \times 32$ &
% 200 &
% RandomCrop(32, pad=4), HorizontalFlip \\

% ImageNet &
% $224 \times 224$ &
% 90 &
% RandomResizedCrop(224), HorizontalFlip \\
% \bottomrule
% \end{tabular}
% \end{table}

\begin{table}[H]
\centering
\caption{\textbf{Overview of experimental variants and controlled modifications from the base StoMPP scheme in Table~\ref{tab:training_config_ablations}.}
Unless otherwise stated, all experiments use the same backbone, data preprocessing, optimizer, batch size, training protocol, and precision policy described in Section~\ref{sec:exp_setup}. Each variant modifies exactly one component of the base scheme.}
\label{tab:experiment_overview}
\small
\begin{tabular}{llll}
\toprule
\textbf{Experiment} & \textbf{Component} & \textbf{Modification} & \textbf{Reference} \\
\midrule
STE baseline & Training rule & STE instead of StoMPP & Tab.~\ref{tab:main_results} \\

StoMPP (STE-free) & -- & Base scheme (layerwise, cubic, $r{=}100$) & Tab.~\ref{tab:main_results} \\

StoMPP+STE & -- & Base scheme (layerwise, cubic, $r{=}100$) & Tab.~\ref{tab:main_results} \\

\midrule
\textbf{Ordering ablation} & Mask ordering &
Layerwise / Global / Reverse & Sec.~\ref{sec:layerwise_ablation} \\

\textbf{Policy ablation} & Mask policy &
Stochastic vs.\ deterministic (BWN only) & Sec.~\ref{sec:layerwise_ablation} \\

\midrule
\textbf{Schedule ablation} & Freezing schedule $p(t)$ &
Cosine / Linear / Quadratic / Cubic / Flipped quad. & Sec.~\ref{sec:hp_ablations} \\

\textbf{Refresh ablation} & Refresh rate $r$ &
$r \in [10, 10^4]$ & Sec.~\ref{sec:hp_ablations} \\

\midrule
\textbf{Hybrid (A/W)} & Training rule split &
StoMPP activations, STE weights & Sec.~\ref{sec:hybrids} \\

\textbf{Hybrid (W/A)} & Training rule split &
StoMPP weights, STE activations & Sec.~\ref{sec:hybrids} \\

\midrule
\textbf{Epoch ablation} & Training length &
Vary total epochs & Sec.~\ref{sec:epoch_ablation} \\

\textbf{LR ablation} & Learning rate &
$\mathrm{lr}\in\{10^{-3},10^{-2},10^{-1}\}$ & Sec.~\ref{sec:lr_ablation} \\

\bottomrule
\end{tabular}
\end{table}

\begin{table}[H]
\centering
\caption{Training configuration for ablations.}
\label{tab:training_config_ablations}
\small
\begin{tabular}{ll}
\toprule
\textbf{Category} & \textbf{Setting} \\
\midrule
Dataset & CIFAR-100 \\
Input resolution & $32 \times 32$ \\
Train batch size & 256 \\
Test batch size & 256 \\
Data augmentation & RandomCrop(32, pad=4), HorizontalFlip \\
Normalization & mean (0.5071, 0.4865, 0.4409), std (0.2673, 0.2564, 0.2762) \\
Model architecture & ResNet-18 \\
Model variant & StoMPP-BNN \\
Optimizer & SGD (Nesterov momentum) \\
Learning rate & 0.1 \\
Momentum & 0.9 \\
Weight decay & 0 \\
LR schedule & Constant \\
Loss & Cross-entropy \\
Label smoothing & 0.0 \\
Masking scheme & Layerwise progressive masking \\
Mask schedule & Cubic ($p: 0 \rightarrow 1$) \\
Mask refresh rate & 100 \\
Activation function (unfrozen) & Clip \\
Frozen activation & Sign \\
Training epochs & 200 \\
Precision & FP32 \\
\bottomrule
\end{tabular}
\end{table}

\section{Training Curves of Variants}
\label{sec:appendix_b}

While we discuss the sawtooth-like training curves in Section~\ref{sec:training_dynamics} and Figure~\ref{fig:hp_sweep_and_curves} (b-d), we find that the hybrid architectures introduced in Section~\ref{sec:hybrids} and Figure~\ref{tab:compare_hybrids}, as well as BWNs, exhibit different training curves than STE or the standard layerwise StoMPP training algorithm.

\subsection{BWN Training Curves}

In the context of BWNs (Figure~\ref{fig:app_bwn_test}, Figure~\ref{fig:app_bwn_train}), we find that StoMPP shows a "swoop", initially increasing as it trains to the task, followed by a plateau or dip, and finally a region that accelerates tot the final accuracy again. We believe this may be because the network is harder to train as it is rapidly binarizing, and that performance and the representational capacity loss are "fighting" in the saddle. As the bulk of weights are trained, the network is mostly binary, and it can effectively learn the final portion of the network. This training curve behavior is of course dependent on the p value throughout training, and suggest further investigation into the nature of these curves.

\begin{figure}[H]
    \centering
    \includegraphics[width=0.9\linewidth]{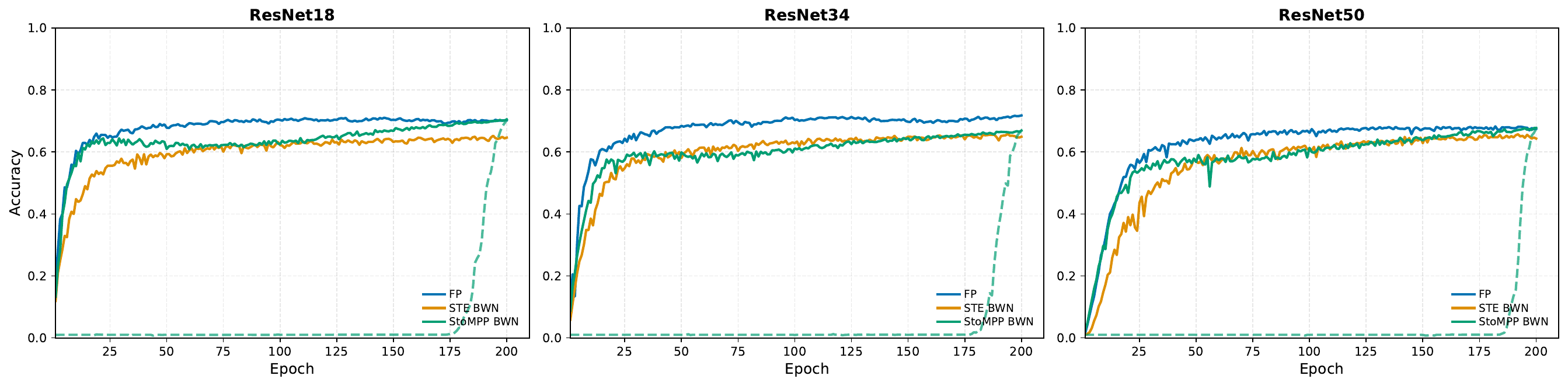}
    \caption{Testing Accuracy Curves for BWNs}
    \label{fig:app_bwn_test}
\end{figure}

\begin{figure}[H]
    \centering
    \includegraphics[width=0.9\linewidth]{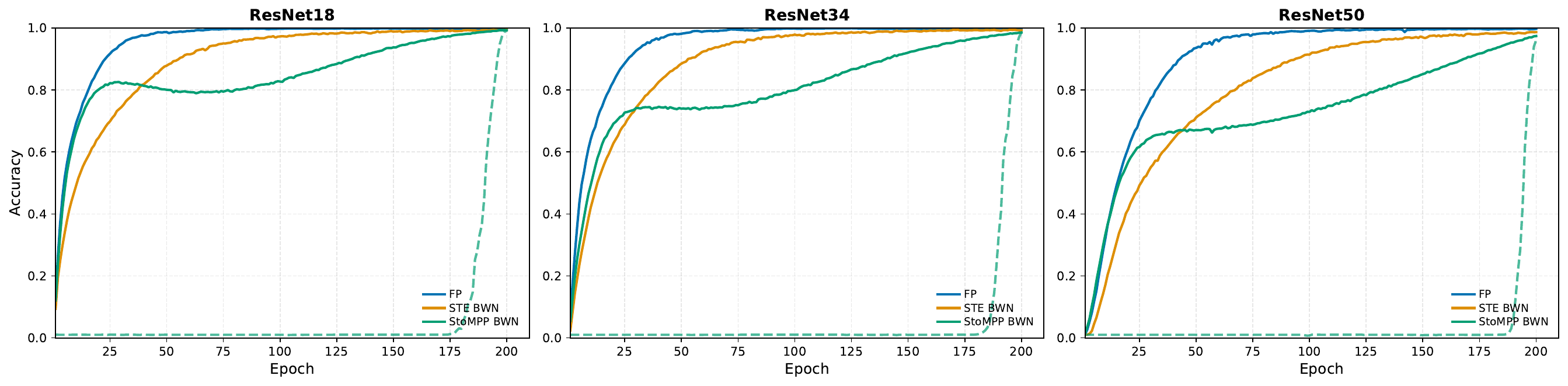}
    \caption{Training Accuracy Curves for BWNs}
    \label{fig:app_bwn_train}
\end{figure}

\subsection{StoMPP BNN Hybrids}

We observe that when Hybrid (StoMPP activations, STE weights) and Reverse Hybrid (StoMPP weights, STE activations) have significantly different training behavior beyond their final results. We find that while StoMPP has the strong sawtooth effect described as it learns each layer. In the context of a layerwise masking scheme, we find that reverse hybrid exhibits a sawtooth curve while hybrid does not. 

\begin{figure}[H]
    \centering
    \includegraphics[width=0.9\linewidth]{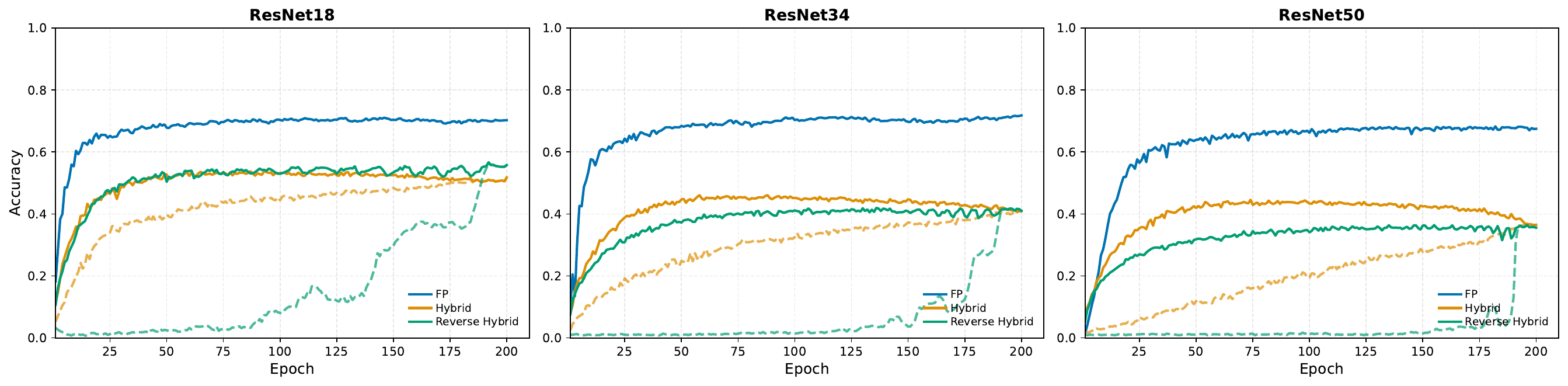}
    \caption{Testing Accuracy Curves StoMPP Hybrids}
    \label{fig:app_bwn_test}
\end{figure}

\begin{figure}[H]
    \centering
    \includegraphics[width=0.9\linewidth]{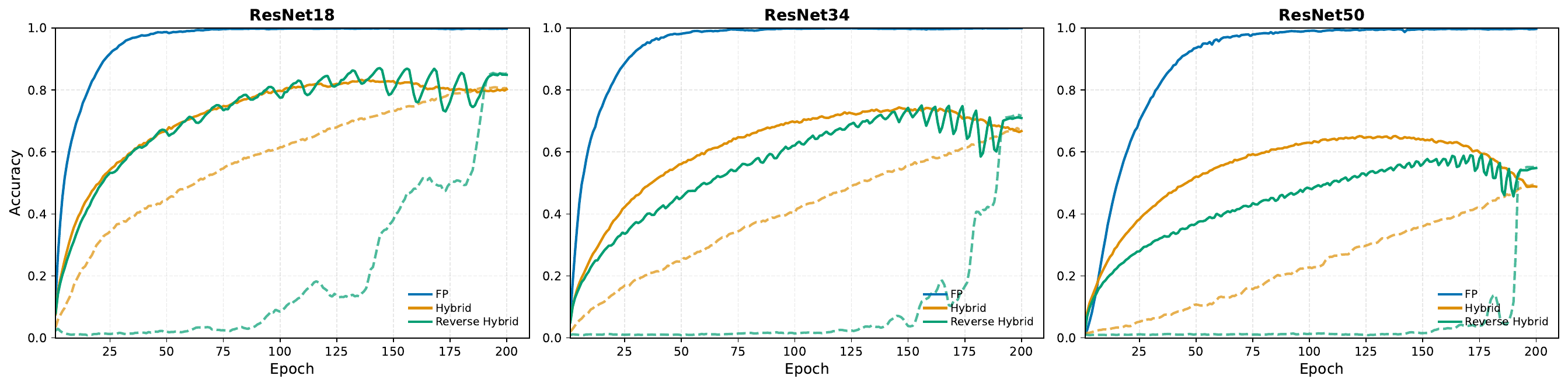}
    \caption{Training Accuracy Curves for StoMPP Hybrids}
    \label{fig:app_bwn_train}
\end{figure}

When applying the hybrid techniques using the global masking technique (which result in worse final performance, as shown in Section~\ref{sec:main_results}), we find that this sawtooth effect disappears. Based on the understanding that a global network mask has a plateau as $p$ rises, it makes sense that a layerwise mask applied this per layer. Despite the absence of a sawtooth effect for global masking, we still find that the convergence behavior of hybrid and reverse hybrid vary significantly. We find that reverse hybrid exhibits the same general contour as the BWN, with increasing accuracy as the freezing finalizes, while hybrid drops as the freezing finalizes. These effects are especially pronounced on the training accuracy since that is what the model is actually learning, although they are visible on the testing accuracy as well.

\begin{figure}[H]
    \centering
    \includegraphics[width=0.9\linewidth]{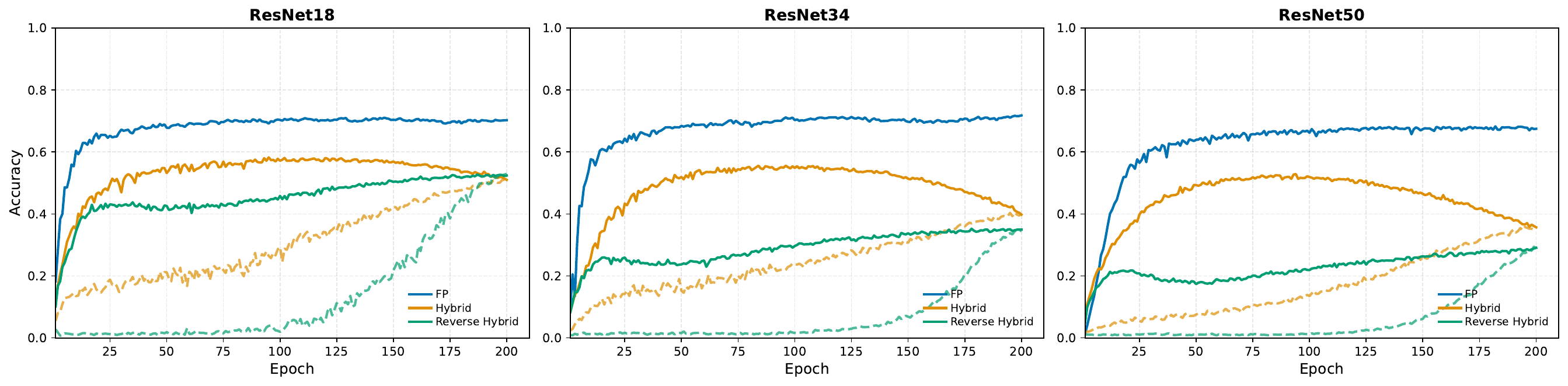}
    \caption{Testing Accuracy Curves for StoMPP Hybrids}
    \label{fig:placeholder}
\end{figure}

\begin{figure}[H]
    \centering
    \includegraphics[width=0.9\linewidth]{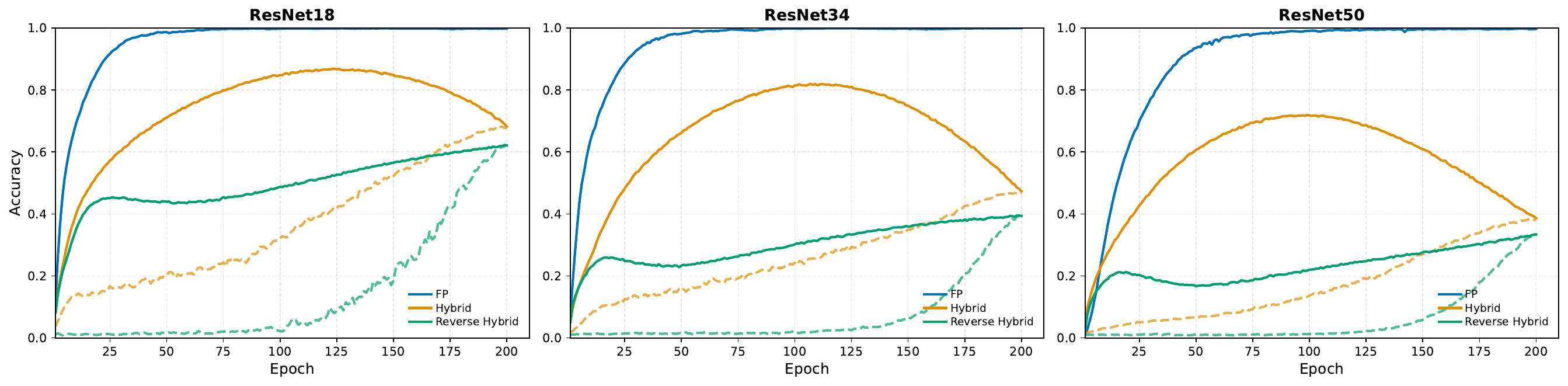}
    \caption{Training Accuracy Curves for StoMPP Hybrids}
    \label{fig:placeholder}
\end{figure}

\newpage
\subsection{Finetuning and BiReal-Net}

The training curves of BiReal-Net also show interesting behavior when combined with StoMPP. When applying layerwise masking in this context, we find that BiReal-Net significantly limits the sawtooth oscillations noted in Section~\ref{sec:training_dynamics}. We believe this is due to the additional residual connections added to the network, adding a signal as the activations and weights from the preceding layers are binarized. We also find that the quantized and unquantized version of StoMPP match much more closely throughout training, rather than just at the end, when training with BiReal-Net. The improved apparent stability of training StoMPP with BiReal network suggests possible architectural improvements may improve the ability of this technique. 

\begin{figure}[H]
    \centering
    \includegraphics[width=0.7\linewidth]{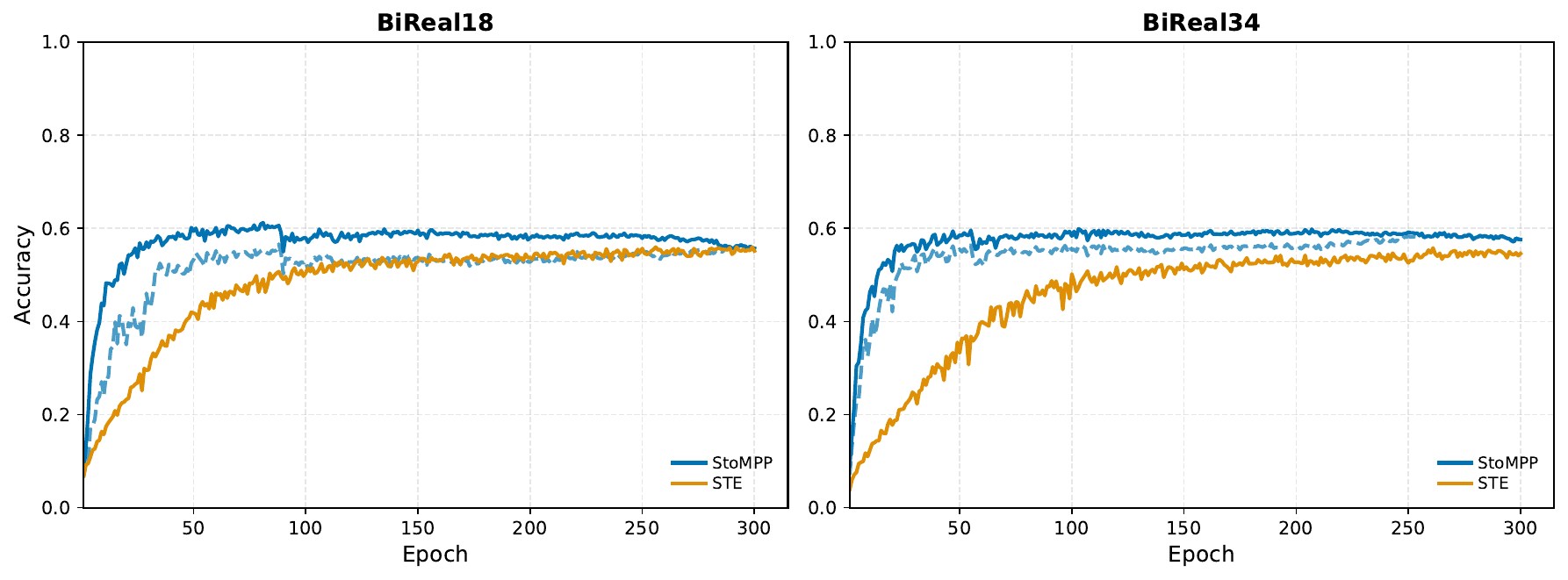}
    \caption{Testing Accuracy Curves for BiReal training from scratch}
    \label{fig:placeholder}
\end{figure}

\begin{figure}[H]
    \centering
    \includegraphics[width=0.7\linewidth]{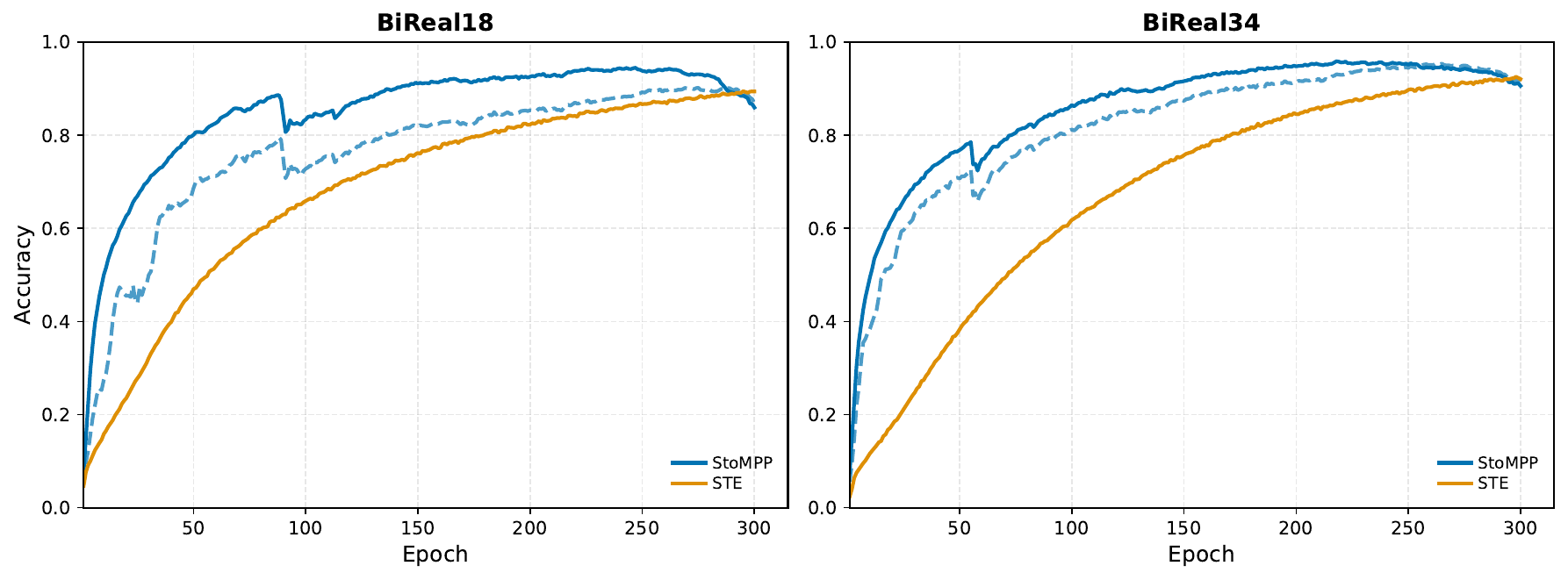}
    \caption{Training Accuracy Curves for BiReal training from scratch}   
    \label{fig:placeholder}
\end{figure}

We find also that training StoMPP/STE from a pretrained network, either in BiReal-Net or for typical ResNets, that both techniques are able to retain their performance. These networks maintain a generally stable result even across freezing with limited "sawtooth" behavior.

\begin{figure}[H]
    \centering
    \includegraphics[width=0.7\linewidth]{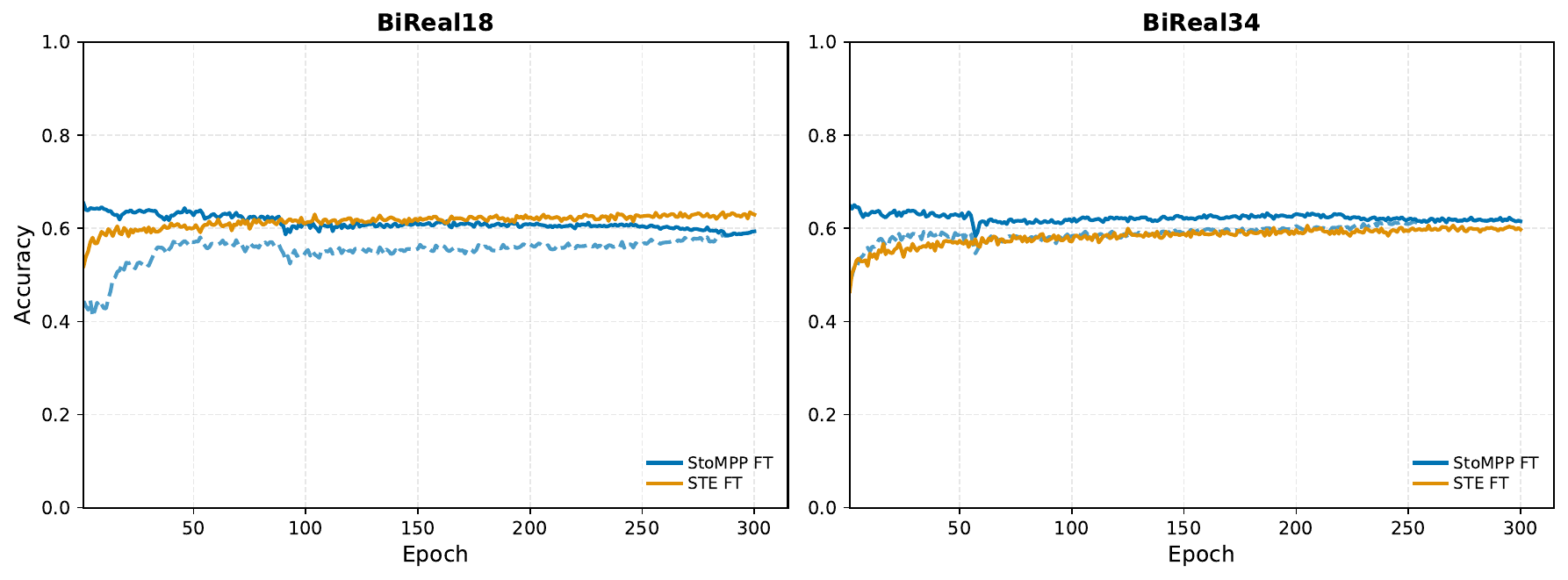}
    \caption{Testing Accuracy Curves for finetuned (FT) BiReal training}
    \label{fig:bireal_ft_test}
\end{figure}

\begin{figure}[H]
    \centering
    \includegraphics[width=0.7\linewidth]{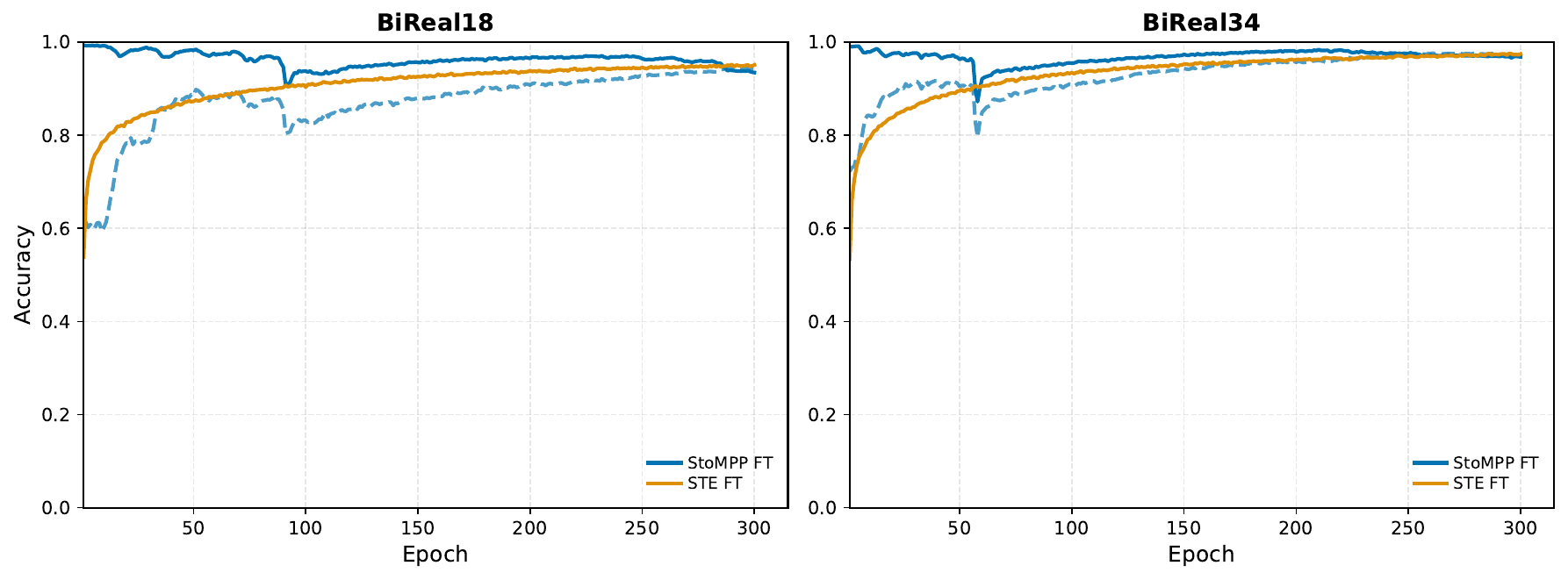}
    \caption{Training Accuracy Curves for finteunted (FT) BiReal training}   
    \label{fig:bireal_ft_train}
\end{figure}
\section{Performance of STE-Free StoMPP / STE Hybrids}
\label{sec:hybrids}

StoMPP and STE act on different points in binary training, so we test whether mixing them can improve performance. We consider two hybrids that swap the training rule used for weights and activations: \textbf{Hybrid (A/W)} applies STE-free StoMPP to \emph{activations} and STE to \emph{weights}, while \textbf{Reverse Hybrid (W/A)} applies STE-free StoMPP to \emph{weights} and STE to \emph{activations}. All methods are trained under the same protocol.

Table~\ref{tab:compare_hybrids} shows that all StoMPP-based variants substantially outperform pure STE across depths. The best choice depends on depth: for R18 and R34, \textbf{Reverse Hybrid (W/A)} attains the highest test accuracy, suggesting that much of the gain comes from improving \emph{weight} optimization while keeping STE for binary activations. For the deeper R50, \textbf{fully StoMPP} achieves the best test accuracy, indicating that as depth increases, applying StoMPP to both weights and activations becomes increasingly important. Applying StoMPP only to activations (Hybrid, A/W) is consistently weaker than the weight-focused variants, reinforcing that weight optimization is the primary driver of gains, while full StoMPP scales most reliably to deeper networks. See Appendix~\ref{sec:appendix_b} for additional observation on different qualitative training-curve behavior in the BWN setting.

\begin{table}[H]
\centering
\caption{Comparison of hybrid techniques on CIFAR-100.
Entries report \textit{Train / Test} accuracy (\%) on quantized networks.
(A/W): StoMPP activations, STE weights. (W/A): StoMPP weights, STE activations.}
\label{tab:compare_hybrids}
\begin{tabular}{lccc}
\toprule
\textbf{Method} & \textbf{R18} & \textbf{R34} & \textbf{R50} \\
\midrule
STE             & 66.6 / 49.1 & 37.7 / 33.7 & 28.8 / 26.7 \\
\midrule
StoMPP          & 83.2 / \textbf{53.8} & 72.4 / 39.8          & 54.6 / \textbf{40.2} \\
Hybrid (A/W)    & 80.5 / 51.8          & 67.6 / 40.9          & 49.3 / 36.4          \\
Rev.\ Hybrid (W/A) & 85.3 / \textbf{55.8} & 71.6 / \textbf{41.0} & 55.6 / 35.6          \\
\bottomrule
\end{tabular}
\vspace{-1em}
\end{table}

\section{Deployment and Training of Binary Neural Network Variants}
\label{sec:app_memory}

\subsection{\textbf{Inference Overhead of Binary Neural Networks.}}

To quantify the performance and efficiency gains of binary weight networks and binary neural networks with StoMPP, STE, amd StoMPP+STE in Table~\ref{tab:app_efficient}. Because StoMPP, STE, and StoMPP+STE all serve as \textit{training time} techniques and result in the same performance at deployment, these numbers follow directly from\textit{ A comprehensive review of binary neural network} by \cite{Yuan_Agaian_2023}.

\begin{table}[H]
\centering
\footnotesize
\caption{Inference efficiency of ResNet-18 variants on CIFAR-100. 
BOPs, FLOPs, and OPs follow directly from \cite{Yuan_Agaian_2023}.}
\label{tab:app_efficient}
\begin{tabular}{llcccccc}
\toprule
\textbf{Setting} & \textbf{Method} & \textbf{Bits (W/A)} & \textbf{BOPs ($\times10^8$)} & \textbf{FLOPs ($\times10^8$)} & \textbf{OPs ($\times10^8$)} & \textbf{Top-1 (\%)} \\
\midrule
Full Precision & ResNet-18 & 32/32 & --- & 18.1 & 18.1 & 71.1 \\
\midrule
\multirow{2}{*}{BWN} & STE   & 1/32 & 1.70 & 1.31 & --- & 64.6 \\
                     & StoMPP (STE-free) & 1/32 & 1.70 & 1.31 & --- & 69.5 \\
                     & StoMPP + STE & 1/32 & 1.70 & 1.31 & --- & 69.5 \\
\midrule
\multirow{2}{*}{BNN} & STE   & 1/1 & 1.70 & 0.36 & 1.67 & 49.1 \\
                     & StoMPP (STE-free) & 1/1 & 1.70 & 0.36 & 1.67 & 53.8 \\
                     & StoMPP + STE & 1/1 & 1.70 & 0.36 & 1.67 & 53.8 \\
\bottomrule
\end{tabular}
\end{table}

\subsection{\textbf{StoMPP Training Overhead.}}

StoMPP is an architecture-agnostic training method. We summarize practical choices for applying it to modern feedforward networks, including which layers to schedule, scheduling granularity, and computational overhead. StoMPP adds minimal overhead: SoftRefresh samples $O(n/r)$ indices per step, and Eq.\eqref{eq:stompp_forward} is elementwise. By the end of training, all scheduled layers satisfy $p(T)=1$, yielding a fully binarized inference network (weights and activations use $\mathrm{sign}$ throughout scheduled layers). Default hyperparameters are provided in Appendix \ref{sec:appendix_a}.

\section{Hyperparameter Sweep of Forward Masking}
\label{sec:app_hp_forward_masking}

While StoMPP as presented in the paper with layerwise masking is quite robust to hyperparameters, we find that while the particular masking scheme is very important for the performance of a network when \textit{global masking} is used. Aside from applying global masking instead of layerwise masking, the approach is identical to that of Figure~\ref{fig:hp_sweep_and_curves} (a), analyzing ResNet18 over 5 schedulers. See Section~\ref{sec:hp_ablations} for more details on the setup, and Appendix~\ref{sec:appendix_b} for exact training scheme specifics.

\begin{figure}[H]
    \centering
    \includegraphics[width=0.6\linewidth]{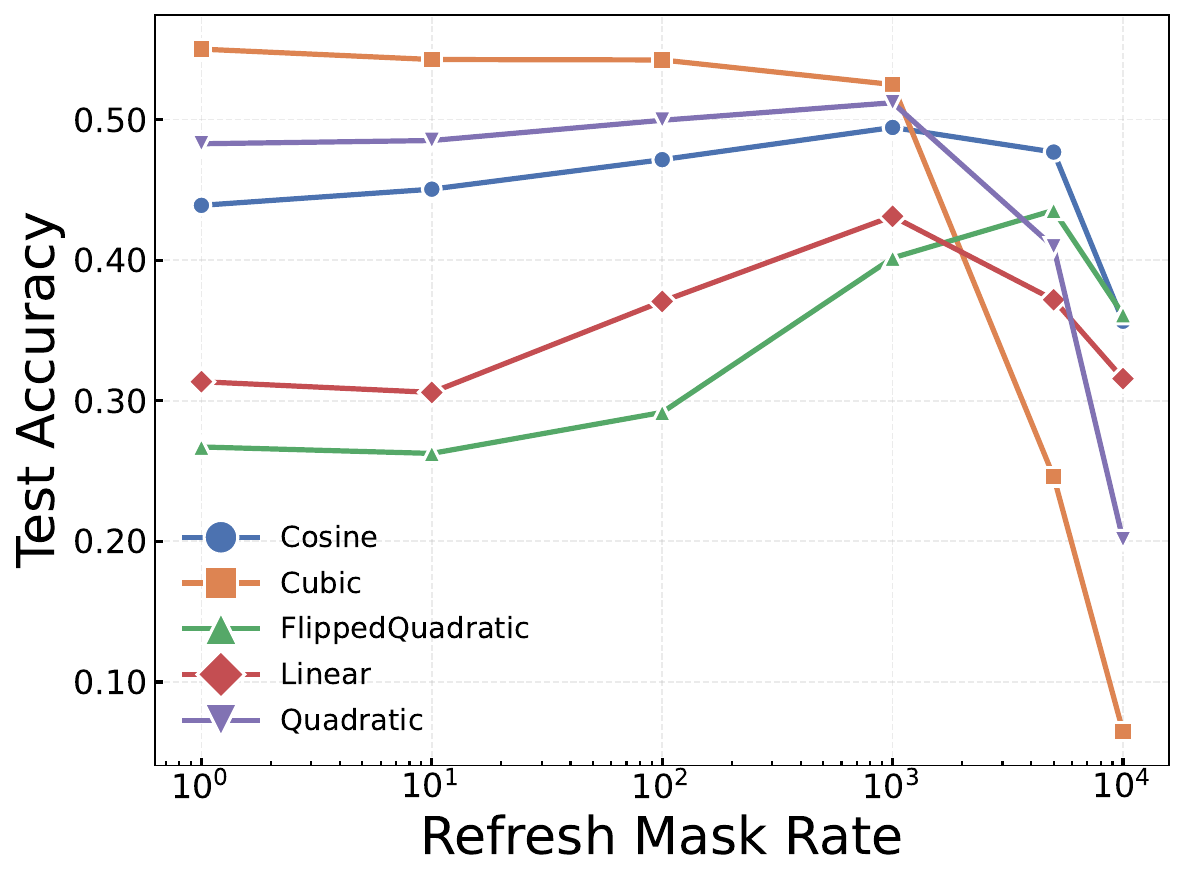}
    \caption{Hyperparameter Sweep of Masking Schemes under Layerwise Masking}
    \label{fig:app_hp_sweep_masking_schemes}
\end{figure}

In a global masking context, the ideal refresh rate varies depending on the schedule quite considerably, and refresh rates that increase more rapidly have increased performance. For example, Cubic > Quadratic > Linear.
\section{Additional Related Work}
\label{sec:app_related_work}

We list additional binary neural network training methods and architectural variants not discussed in the main paper due to space constraints, acknowledging varied approaches and progress in extreme quantization.

\paragraph{Multiple Binary Bases and Approximation Methods.}
ABC-Net \citep{lin2017towards}, AdaBin \citep{tu2022adabinimprovingbinaryneural}, Projection CNN \citep{gu2018projectionconvolutionalneuralnetworks}.

\paragraph{Architectural Improvements.}
MeliusNet \citep{bethge2021meliusnet}, Group-Net \citep{zhuang2022structuredbinaryneuralnetworks}, BinaryDenseNet \citep{Bethge2019BinaryDenseNetDA}, Circulant Binary CNN \citep{liu2019circulantbinaryconvolutionalnetworks}.

\paragraph{Activation Function Design.}
ReActNet (PReLU-based) \citep{liu2020reactnet}, Unbalanced Activation \citep{kim2021improving}, Regularizing Activation Distributions \citep{ding2019regularizingactivationdistributiontraining}.

\paragraph{Optimizer and Training Strategies.}
AdamBNN \citep{liu2021adam}, Binary Optimizer (BOP) \citep{helwegen2019latentweightsexistrethinking}, SGDAT \citep{sgdat2023}.

\paragraph{Knowledge Distillation.}
Training with Knowledge Transfer \citep{leroux2020training}, Quantization-aware Knowledge Distillation (QKD) \citep{QKD_Kim_2019}, KDG-BNN \cite{gao2022memristive}.

\paragraph{Loss Functions and Regularization.}
Loss-Aware Binarization \citep{hou2017loss}, BinaryDuo \citep{kim2020binaryduoreducinggradientmismatch}, Defensive Quantization \citep{lin2019defensivequantizationefficiencymeets}.

\section{Computational Resources}
\label{sec:app_compute}

All experiments were conducted on a university high-performance computing cluster using NVIDIA L40S, A100, and H100 GPUs. CIFAR-10 and CIFAR-100 experiments required a few hours per run; ImageNet experiments required approximately 2–3 days per run. The paper includes approximately 117 total runs: ~15 on ImageNet and ~102 on CIFAR-10/100 or equivalent-scale benchmarks (MobileNetV2 on CIFAR-100, BERT on SST-2, BiReal-Net on CIFAR-100). All runs are single-seed. Additional preliminary experiments were conducted during method development on the same hardware at comparable per-run cost.

% \section{Technical appendices and supplementary material}
% Technical appendices with additional results, figures, graphs, and proofs may be submitted with the paper submission before the full submission deadline (see above). You can upload a ZIP file for videos or code, but do not upload a separate PDF file for the appendix. There is no page limit for the technical appendices. 

% Note: Think of the appendix as ``optional reading'' for reviewers. The paper must be able to stand alone without the appendix; for example, adding critical experiments that support the main claims to an appendix is inappropriate. 

%%%%%%%%%%%%%%%%%%%%%%%%%%%%%%%%%%%%%%%%%%%%%%%%%%%%%%%%%%%%

% \newpage
% \input{checklist.tex}

\end{document}